\documentclass{article}
\PassOptionsToPackage{numbers,sort}{natbib}
\usepackage{style}

\usepackage{natbib}
\usepackage[utf8]{inputenc} 
\usepackage[T1]{fontenc}    
\usepackage{url}            
\usepackage{booktabs}       
\usepackage{amsfonts}       
\usepackage{nicefrac}       
\usepackage{microtype}      
\usepackage{xcolor}         
\usepackage{stfloats}

\usepackage{times}
\usepackage{soul}
\usepackage[hidelinks]{hyperref}
\usepackage[small]{caption}
\usepackage{graphicx}
\usepackage{amsmath}
\usepackage{amsthm}
\usepackage{algorithm}
\usepackage{algorithmic}
\usepackage{enumerate}
\usepackage{mdwlist}
\usepackage{amssymb}
\usepackage{paralist}
\usepackage{placeins}
    
\usepackage{xspace}    
\usepackage{paralist}
\usepackage{url}

\newcommand{\eg}{e.\,g.,\xspace}
\newcommand{\ie}{i.\,e.,\xspace}
\newcommand{\cf}{cf.\xspace}
\newcommand{\Eg}{E.\,g.,\xspace}
\newcommand{\Ie}{I.\,e.,\xspace}

\newcommand{\andor}{\textsc{ANDOR}\xspace}

\newcommand{\eat}[1]{}

\title{Saliency Methods are Encoders:\\ Analysing Logical Relations Towards Interpretation}

\iftrue
\author{
Leonid Schwenke$^1$
\and
Martin Atzmueller$^{1,2}$
\affiliations
$^1$Semantic Information Systems Group (SIS)\\
$^2$German Research Center for Artificial Intelligence (DFKI)
\emails
\{leonid.schwenke, martin.atzmueller\}@uni-osnabrueck.de
}
\fi

\begin{document}

\maketitle

\begin{abstract}
    With their increase in performance, neural network architectures also become more complex, necessitating  explainability. Therefore, many new and improved methods are currently emerging, which often generate so-called saliency maps in order to improve interpretability. Those methods are often evaluated by visual expectations, yet this typically leads towards a confirmation bias. Due to a lack of a general metric for explanation quality, non-accessible ground truth data about the model's reasoning and the large amount of involved assumptions, multiple works claim to find flaws in those methods. However, this often leads to unfair comparison metrics. Additionally, the complexity of most datasets (mostly images or text) is often so high, that approximating all possible explanations is not feasible. For those reasons, this paper introduces a test for saliency map evaluation: proposing controlled experiments based on all possible model reasonings over multiple simple logical datasets. Using the contained logical relationships, we aim to understand how different saliency methods treat information in different class discriminative scenarios (\eg via complementary and redundant information). By introducing multiple new metrics, we analyse propositional logical patterns towards a non-informative attribution score baseline to find deviations of typical expectations. Our results show that saliency methods can encode classification relevant information into the ordering of saliency scores.
\end{abstract}

\section{Introduction}
While deep learning (DL) is getting more powerful, understandability is still lagging behind~\citep{ruping2006learning}. With a better understanding of the respective model, debugging, finding unwanted biases and increasing safety for critical environments is simpler and more cost-efficient~\citep{molnar2020interpretable}. In the context of eXplainable Artificial Intelligence (XAI), saliency maps are a typical type of explanation, which assign attribution scores in order to form input-based heatmaps.
The saliency scores show a certain relevancy of specific input values towards the output, relative to all other values of the current local input. While these saliency maps have already successfully been used for debugging, \eg the Clever Hans problem~\citep{lapuschkin2019unmasking}, the evaluation of the quality of saliency methods is still controversial~\citep{adebayo2018sanity, ju2021logic}. For evaluating \eg image data, the highlighted area is often compared with the area expected by a human. This, however, often leads to a confirmation bias~\citep{adebayo2018sanity}, \eg demonstrated by adversarial attacks~\citep{eykholt2018robust}. Thus, this is not sufficient for verifying the quality of a saliency map. 
We tackle this bias by approximating possible model reasonings using ground truth data. For this purpose, we introduce a logical dataset-framework \andor, while also considering the statistical training set distributions.

Due to the locality in computer vision (CV) and time series tasks, often close inputs share similar information. Additionally, due to the complexity of most tasks and the incomplete nature of the datasets, it is often not possible or feasible to capture all possible ground truth model reasoning data $D$, \ie which information decides the model output. However, by fully grasping which input contains which information, all possible ground truth reasoning scenarios $R$ can be derived. If a DL-model $f$ reaches a full understanding of the task, $f$ must follow one $r \in R$ and thus can be evaluated.
Given a set of typical expectations for saliency maps and within a controlled experiment setting, in this paper we analyse multiple simple propositional logical datasets towards if and how well different saliency methods capture a possible explanation $R$ in different logical operations (AND, OR, XOR). By using this simple logical setting as our framework, we aim to better understand fundamental information handling principles of different saliency methods and enable a basic trust test. \Ie if problems in simple relations exist, then they can also occur in more complex ones (trust).

\paragraph{Summary of our contributions}
\begin{enumerate}
    \item We create a simple logical dataset framework \andor for analysing a set of saliency methods regarding their global information handling.
    \item Using typical expectations on saliency methods, we define new metrics to enable a trust test. Using those, we evaluate multiple saliency methods towards our extracted possible reasoning $R$ for multiple \andor datasets, which all contain an irrelevant saliency score baseline.
    \item We show that saliency methods can encode relevant classification information into the order of saliency scores, and discuss possible reasons for this in detail.
\end{enumerate}

\section{Related Work}

Multiple approaches exist which try to find flaws in saliency methods~\citep{ju2021logic, adebayo2018sanity, kindermans2019reliability, shah2021input, kokhlikyan2021investigating}. However, it is hard to find a common ground on which method is actually the best.
\citet{adebayo2018sanity} and \citet{sixt2020explanations}, for example, showed that multiple saliency methods are not input sensitive and approximate somewhat general edge detectors. \citet{li2021experimental} summarizes a few qualitative metrics, which typically try to capture abstract properties, due to the general lack of model reasoning ground truth; \ie a valid reference for comparison is missing as the model internal decision-making are unknown. However, while multiple desired properties and metrics for explanations exist~\citep{rojat2104explainable, carvalho2019machine}, as well as multiple approaches to find flaws, \citet{ju2021logic} showed that most approaches for qualitative evaluation run into logical reasoning traps. 
Here, we argue this is caused by the fact that for most approaches image or text data is used, for which all possible reasoning ground truths are very hard to grasp. While the type of data is essential for the efficiency of applying different saliency methods~\citep{kokhlikyan2021investigating}, we argue, that by understanding and finding basic flaws we can improve and hence increase the trustworthiness of those methods. \Eg multiple valid ways exist to solve a specific task, even if it does not seem plausible~\citep{jacovi2020towards}. Therefore, it is hard to verify the true reasoning (faithfulness) of a model, \ie to fully understand it and extract comprehensive ground truth reasoning data.

\citet{rong2022consistent} showed that masks can contain classification relevant information. They demonstrated this on image data, where shapes and redundant information are very common, but did not fully consider all possible redundancies. While considering all redundant information, we on the other hand show that different saliency methods use mask/order information to encode information in a more complex way than \eg shapes indicated by mask contours in images.

Using logical operations or reasoning for DL interpretation is not completely new~\citep{zhang2021survey, marques2023logic, fan2017revisit}, sometimes also applied as approximation on symbolic networks \citep{garcez2022neural, srinivasan2019logical}. \citet{zhang2021survey} and \citet{pedreschi2019meaningful} even highlight the need of logical reasoning for better explanations, thus showing the advantage to approximate logical reasoning with attribution techniques.
\citep{yalcin2021evaluating} tried to verify the necessity of certain inputs in a logical formula dataset using attribution scores of non-DL methods. However, they only captured the dataset ground truth and failed to capture the model reasoning, due to suboptimal model performance, not using all possible input combinations and neglecting statistical structures. Further, \citep{tritscher2020evaluation} used a logical dataset with a 100\% accurate DNN to verify saliency scores; however, they only approximated the maximal information coverage and thus mistreated redundant information, compared to \citep{yalcin2021evaluating}. We are considering those flaws using a DNN, by including a relative non-information attribution score baseline, by differentiating between maximal and minimal information coverage and by making an even more exhausting analysis using more metrics; while also focusing fully on the logical information characteristics of the logical framework (\andor) we introduce and approximate all possible model reasoning. Hence, our work exceeds the scope of~\citet{tritscher2020evaluation} and \citet{yalcin2021evaluating} considerably.

\section{Experiment Setup and Assumptions}
In this section, we introduce necessary formal notation, define our \andor dataset and discuss multiple assumptions/expectations on saliency maps. Based on that, we define multiple metrics and describe our experimental setup.

\subsection{Information Flow}
\label{sec:informationflow}

In the following, we introduce formal notation for our task.
We consider a function $f\colon D \rightarrow C$ (representing a classification task $T$), given a finite set of classes $C$. Let $D$ denote a dataset that captures all possible instantiations with respect to a given feature set. Thus, $D$ contains all possible data samples $d\in D$ of length $l$, with $d=(d_1, \dots, d_l)$ being a tuple of input (features). An input $d_j \in d$ at position $j$ is thus taken from the finite (universal) input domain $M$, \ie $d_j\in M$ and $D = M^l$. 
For deriving/explaining a class $c \in C$ for a specific $d \in D$ (with $f(d) = c$), however, often only a subset of inputs $s \subseteq d$ is required.
A class $c \in C$ is a set of hidden information $c$ represented by sets $c_i \subseteq H$, considering a universal set $H$ of all hidden class discriminative information of any form, \eg also dataset distribution or implicit information; \ie each $c_i$ represents a distinct valid way to derive class $c$. Because each dataset can have unique class discriminative decisions to derive a class, we combined all possible decision information under the term hidden information; \eg a specific decision tree or a logical formula indicates a possible decision processes $c_i$. However, this does not mean that no other valid decision criteria exists.
In our context, this means that a function $f^D$ for $T$ exists that can extract a set of hidden information for each input $d_j \in d, d \in D$. We define this as $f^D\colon D \times \{1, \dots, l\} \rightarrow Q$, with $Q \subseteq H$. Modelling all $q \in f^D(d, j)$ however is often quite hard, because this relates to knowing all possible ways to derive a class. The function $f^D$ is subject to two important constraints: (1) $\forall d \in D\colon (f(d) = c^1) \rightarrow \exists c_i \in c^1\colon c_i \subseteq \bigcup\limits_{d_j \in d} f^D(d,j)$, \ie the hidden information indicated by the union of all $d_j \in d$ must contain the needed information for at least one $c_i \in c^1$. (2) Each $d \in D$ contains only the information set of exactly one class:  $\forall c_j \in c^2 \in C\colon c_j \subseteq \bigcup\limits_{d_j \in d} f^D(d,j) \rightarrow c^1 = c^2$. With $f^D$ we can derive a set of all $s \subseteq d$ that contain the information needed for $f(d) = c$, defined as $R^d = \{s \mid \exists c_i \in f(d)\colon c_i \subseteq \bigcup\limits_{d_j \in s} f^D(d,j) \wedge s \subseteq d\}$, assuming $s$ keeps the position information of $d$. $R_{min}^d = \{r_1 \mid r_1 \in R^d \wedge \neg \exists r_2 \in R^d\colon |r_2| < |r_1|\}$ is the minimal information coverage for input $d$, \ie all minimal sets of relevant features to derive class $f(d) = c$. $R_{max}^d = \{r \mid r\in R^d \wedge \forall d_j \in r \colon \exists c_i \in f(d) \colon f(d, j) \cap\ c_i \neq \varnothing \}$ is the maximal information coverage of $d$, \ie including all relevant inputs that contain class relevant information. All other inputs are irrelevant for class $f(d) = c$. Those definitions describe our goal, \ie to find a complete estimation on $R^d_{min}$ and $R_{max}^d$ (all possible way to derive a class) for the simple dataset \andor, as described below. Thus, it is possible to check if the suggested saliency score ranking/ordering matches one possible reasoning $r\in R^d_{min}$. For this, we later introduce a non-informative input baseline, to differentiate between relevant and non-relevant inputs.

\begin{figure*}[!b]
	\centering
	\includegraphics[width=1.6\columnwidth]{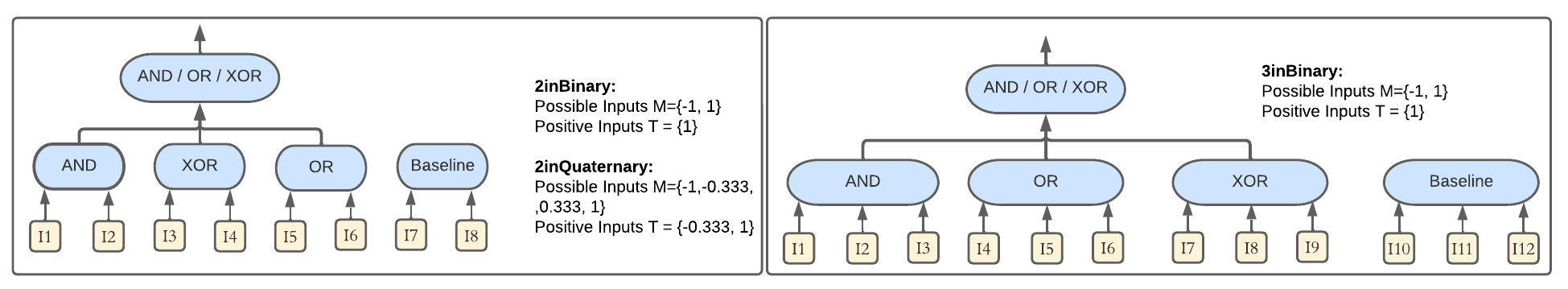}
	\caption{Depicting the three different \andor test-parameter-instances for our experiments. Resulting in datasets of sizes $2^8=256$ (2inBinary), $4^8=65.536$ (2inQuaternary) and $2^{12}=4.096$ (3inBinary), because we take all possible inputs.}
	\label{fig:dataset}
\end{figure*} 

\subsection{Assumptions on Saliency Maps}
\label{sec:assumptions}
As interpretability and explainability are quite vaguely defined, \cf~\citet{zhang2021survey}, assumptions for saliency maps are either not very clearly defined as well or only describe partially what they actually mean~\citep{zhang2021survey}.
With many unclear assumptions, the question arises, how to actually verify the quality of a saliency map? Often, models are evaluated on an ablation of inputs (often using masks), to quantify the ranking between inputs~\citep{hooker2019benchmark, shah2021input, rong2022evaluating,sturmfels2020visualizing}. Thus, based on those evaluation procedures and the logical information flow (see Section~\ref{sec:informationflow}), we consider the following basic assumptions/expectations:

\begin{enumerate}[(A)]
    \item The ranking order between scores is relevant, \ie an input with a higher saliency score than another input is more relevant in the model's local classification process. Therefore, a lower scored input has less, no, or even class contradicting information (w.r.t the model's decisions).
    \item If two inputs contain redundant information, at least one input needs to be relevant (\ie as described above).
    \item The saliency score ranking should not depend on independent inputs that do not contribute to the output.
    \item Dataset distributions and implicit information (\eg shape of masked inputs) can contribute towards information which is relevant for the class.
\end{enumerate}
Based on those assumptions above, we formulate additional, more specific assumptions for our task: 
\begin{enumerate}[(A)]
    \item[(E)] Class irrelevant inputs $d_j \in d,\, d \in D$ act as (non-informative) input baseline, with $\forall r \in R^d_{max} \colon d_j \not\in r$. Hence, such a baseline input $d_j$ should not have a higher saliency score than any class relevant input $d_k \in r \in R^d_{min}$ (see Assumptions A, B and C).
    \item[(F)] If the saliency metric approximates classification information (Assumptions A, B and C), for a logical dataset, then the logical accuracy should be preserved as long as possible due to the input independence and clear $R_{min}$.
\end{enumerate}

\subsection{Dataset}
\begin{figure}[ht!]
	\centering
	\includegraphics[width=0.99\columnwidth]{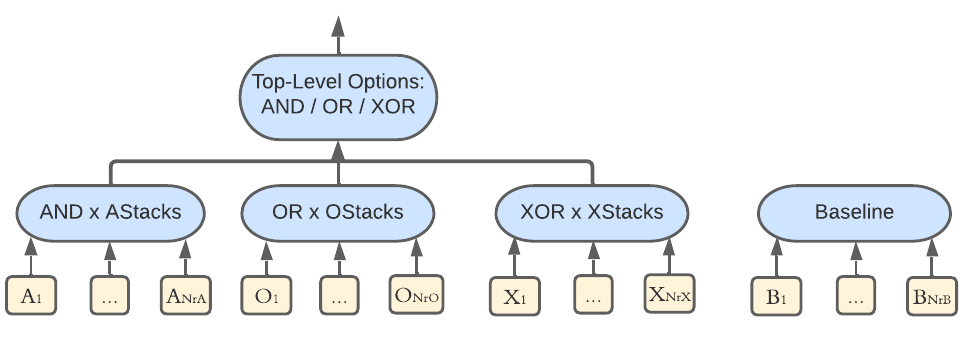}
	\caption{Framework for the \andor dataset.}
	\label{fig:andor}
\end{figure}

To narrow down our expectations on valid saliency maps, we construct the dataset-framework \andor (provided in our code\footnote{\url{https://github.com/lschwenke/SaliencyMapsAreEncoder}}). An \andor dataset is based on propositional logical operators, modelling a two layered relation between operators. The first layer of the logical formula is described by four different blocks (representing \textit{AND}, \textit{OR}, \textit{XOR} and \textit{Baseline}), where each block contains individually many logical gates (\textit{AStacks, OStacks, XStacks}), for which each gate has a fixed predefined length per block (\textit{NrA, NrO, NrX, NrB}). \textit{XOR} is hereby defined as \emph{true} when having exactly one \emph{true} (positive) input. The baseline block acts as reference for non-informative inputs, \ie relevant inputs should be higher scored than the scores that the baseline receives (Assumption E). Each gate input takes on a value from the domain $M$, limiting all possible input values. Here, a binary case with $M =\{0,1\}$ is the simplest one. To be able to evaluate each binary logical operation, a set of positive representations $T \subseteq M$ is given, where all values $m \in M$ with $m \not \in T$ are handled as a negative gate inputs. The final output is decided by the respective \textit{AND}, \textit{OR} or \textit{XOR} gate (top-level). While non-binary outputs would be possible, for simplicity and clarity we only focus on the binary outputs (True/False or 1/0), which can naturally occur in more complex sub-settings. Figure \ref{fig:andor} illustrates this concept. To analyse different situations we deploy 9 different \andor datasets for our experiments, based on the number of three different top-levels multiplied by our three parameter settings: (\textit{2inBinary}) 2 inputs per gate, limited to 1 stack per gate-type, with M=\{-1, 1\} and T=\{1\}; (\textit{2inQuaternary}) similar to 2inBinary but with  M=\{-1, -0.333, 0.333, 1\} and T=\{-0.333, 1\}; (\textit{3inBinary}) similar to 2inBinary but with 3 inputs per gate. A structured example for the three different settings is given in Figure \ref{fig:dataset}. With \andor we aim to analyse fundamental behaviour of saliency maps on redundant, complimentary and exclusive information. To evaluate an \textit{AND} with a positive output, for example, all inputs are needed and provide complementary information; for an \textit{OR} with a positive output, all inputs can provide redundant information, thus only one positive input is needed. We argue those types of relations between inputs will naturally occur in more complex datasets, making \andor a good test dataset due to the fully independent inputs and clear information flow. Hence, this also enables the ability to generate all possible model reasonings, which is not feasible on typical complex datasets. 

\subsection{Expectations on \andor}

To make sure the model understands the task completely, we generate all possible input combinations for each \andor dataset. Hence, to reach an  accuracy of 100\%, the model needs to understand which input values per sample are relevant for each class. To minimize statistical biases (class imbalance), we balance the training set via oversampling using \emph{Imbalanced-learn}~\citep{JMLR:v18:16-365}. 
With this and due to the independence of the inputs, we minimize the effects from Assumption~$D$ and can better approximate $R^d_{min}$ per $d \in D$. By comparing all $r \in R^d_{min}$ towards the \textit{Baseline} over the whole dataset, we can find violations of Assumption~$E$. While we only look at binary outputs and in some cases binary inputs, a good explanation technique should also work for simple scenarios, also because \eg complex inputs/outputs can be sometimes clustered into binary class ranges.

The following specifies what information is expected for each logical gate:
\begin{itemize}
    \item \textbf{AND:} For $f(And_{pos}) = 1$, all values are important/complimentary $R_{min}^{And_{pos}}=\{And_{pos}\}$. For $f(And_{neg}) = 0$, only one negative input is enough (redundancy) $R_{min}^{And_{neg}}=\{d_j \mid d_j \in And_{neg} \wedge d_j \not\in T\}$.
    \item \textbf{OR:}  For $f(Or_{neg}) = 0$, all values are important/complimentary $R_{min}^{Or_{neg}}=\{Or_{neg}\}$. For $f(Or_{pos}) = 1$ , only one positive input is enough (redundancy) $R_{min}^{Or_{pos}}=\{d_j \mid d_J \in Or_{pos}\wedge d_J \in T\}$. 
    \item \textbf{XOR:} For $f(Xor_{pos}) = 1$, all values are important (exclusive information) $R_{min}^{Xor_{pos}}={Xor_{pos}}$. For a $f(Xor_{neg}) = 0$, either all inputs are important and negative or two positive inputs are important; making this case more complex $R_{min}^{Xor_{neg}}=\{Xor_{neg} \mid \forall d_j \in Xor_{neg}: d_J = 0\} \cup \{\{d_i, d_j\} \mid\ \exists d_i, d_j \in Xor_{neg} : d_i \neq d_j \wedge d_i = 1 \wedge d_j = 1\}$. 
\end{itemize}

\subsection{Models}
We use two state-of-the-art architectures with multiple parameters towards enhancing generalizability. We apply shallow networks as they perform better on sequential tasks~\citep{wen2022transformers}, also to reduce biases \citep{haug2021baselines}.
Because multiple saliency maps like \eg GradCam~\citep{selvaraju2017grad} are mainly developed for CNNs, we use two ResNet-Blocks~\citep{he2016deep} to construct a CNN-model. Additionally, we also look into a two-layered Transformer~\citep{vaswani2017attention} to also cover Attention-based methods like \eg the Attention enhanced LRP from~\citep{chefer2021transformer}. We test out all possible saliency methods per model, while exploring the effects of certain logical structures. For details about the architectures and hyperparameters, we refer to our open source code\footnote{\label{foo:code}\url{https://github.com/lschwenke/SaliencyMapsAreEncoder}}.

\subsection{Saliency Maps}
In our experiments, we compare 12 different saliency methods shown in Table~\ref{tab:methods}. In the implementation from~\citet{chefer2021transformer} a CLS-Token is used to reduce the saliency score to one per input. Alternatively, we also apply the sum-operation per row, to better capture the attention distribution.

\begin{table}[ht!]
\scriptsize
\centering
\caption{List of all applied saliency methods, while listing appliable models and the implementation source.}
\label{tab:methods}
\begin{tabular}{l|c|c}
\toprule
\multicolumn{1}{c|}{\textbf{Method}} & \multicolumn{1}{c|}{\textbf{Models}} & \multicolumn{1}{c}{\textbf{Implementation}} \\ \midrule
LRP-Full \citep{bach2015pixel} & Both & \citep{chefer2021transformer} \\
LRP-Rollout \citep{abnar2020quantifying} & Transformer & \citep{chefer2021transformer}  \\
LRP-Transformer \citep{chefer2021transformer} & Transformer &  Adapted from \citep{chefer2021transformer}  \\
LRP-Transformer CLS \citep{chefer2021transformer} & Transformer & \citep{chefer2021transformer}  \\
IntegratedGradients \citep{sundararajan2017axiomatic} & Both & Captum \citep{kokhlikyan2020captum} \\
DeepLift \citep{shrikumar2017learning} & Both & Captum \citep{kokhlikyan2020captum} \\
Deconvolution \citep{zeiler2014visualizing} & CNN & Captum \citep{kokhlikyan2020captum} \\
GradCam \citep{selvaraju2017grad} & Both & pytorch-grad-cam \citep{jacobgilpytorchcam} \\
GuidedGradCam \citep{selvaraju2017grad} & Both & Captum \citep{kokhlikyan2020captum} \\
GradCam++ \citep{chattopadhay2018grad} & Both & pytorch-grad-cam \citep{jacobgilpytorchcam} \\
KernelSHAP \citep{lundberg2017unified} & Both & Captum \citep{kokhlikyan2020captum} \\
FeaturePermutation \citep{molnar2020interpretable} & Both & Captum \citep{kokhlikyan2020captum}\\
\bottomrule
\end{tabular}

\end{table}

\subsection{Experimental Setup}
For each parameter combination, we perform an experiment with a 5 fold-cross-validation (for the validation set). Afterwards, a validation model (ROAR \citep{hooker2019benchmark}) is retrained for each saliency method with masked inputs based on one of four thresholds. The two typical approaches for masking are MoRF (Most Relevant First) and LeRF (Least Relevant First) \citep{tomsett2020sanity}. Because MoRF would not necessarily remove redundant information, we only considered LeRF, \ie all high scored values are relevant for the task. Our first threshold is the highest \textit{Baseline} input score per sample, \ie Assumption E. In contrast to~\citet{hooker2019benchmark} we take the average saliency score per sample times a factor, to enable more dynamic masking for samples where more or fewer inputs can be relevant. The three remaining thresholds t1.0, t0.8, t0.5 stand for the used factors, \ie t1.0 = avg. $\times$ 1. For each trained base model, Random Forest Model \citep{breiman2001random} is trained for comparison. Afterwards, all our metrics are calculated. In total, we analysed 144 experiments, resulting in 33600 trained neural network models (including re-trained models). Those result from the 9 dataset settings, the 8 models (4 different configurations $\times$ 2 model types) and 2 different types of train/test splits. We either use an 8/2 split (9/1 for 2inBinary due to small dataset size), or we utilize all data as training, validation and test set, thus making sure that the model has seen and potentially understood every input (Split Test vs Not Split Test). We focus on the split test set, and include the non-split test set for complementing the analysis. For both, the training data is always class balanced. To make sure the model has learned the task without a bias, we primarily consider results where the base model reached 100\% acc. on the split test. This includes 233 base models (considering folds).
The hyperparameters are selected over sample based manual optimization per dataset. For details about the pipeline, we refer towards our code\footnotemark[2]. We ran our experiments for about 30 days on a cluster with 6 nodes, each having an NVIDIA A100, 40GB, 2x AMD EPYC™ 7452 and 512GB of RAM. The intermediate results were about \textasciitilde 1.7TB and the final one about \textasciitilde 3GB.

\subsection{Metrics}
For identifying violations of our assumptions described in Section~\ref{sec:assumptions}, we define the following metrics:

\begin{compactenum}[(1)]
    \item Needed Information below Baseline (NIB): Percentage of samples $d \in D$, where at least one input $r_j \in r, r \in R^d_{min}$ is below the highest \textit{Baseline} input per $d$. This metric checks for Assumption E. If $\mathit{NIB} > 0$, then $R^d_{max}$ does also not meet the assumption as well.
    \item Logical Accuracy: The accuracy after masking the data, by using known logical truth tables, \ie combinations of undefined inputs result in undefined. With this, we check for Assumption F.
    \item Logical Statistical Accuracy: The accuracy after masking the data, by using known logical truth tables and considering probabilities for masked inputs assignments. Therefore, taking \eg an \textit{AND}-gate, if two or more inputs are masked, then the output is more likely to be false.
    \item Full Double Class Assignments (Full-DCA): Similar to truth tables, two equal sets of inputs (ignoring irrelevant inputs) should not lead towards different results. For all samples of the test set, where the original and the retrained model output the same class: Given a threshold $t$, the count where the relevant inputs $\{ \{d_1, \dots, d_{l-NrB}\} \subseteq d \mid d \in D\}$ map towards different classes in the retrained model, \ie the decision relevant information is in the \textit{Baseline} inputs (\cf Assumption C). Considering our Assumptions, as long as relevant inputs can remain (after applying the threshold), the DCA should be 0. To compare multiple settings, we calculate the Full-DCA in percent.
    \item Minimal-DCA: This metric is similar to Full-DCA, but we compare logical gate inputs, where it is clear when this logic gate is relevant for the output class ($|R^d_{min}| = 1$). For \eg the \textit{AND}-top-level comparison per gate: The gate inputs where only this specific gate evaluates negative, to the gate inputs, where the model output is positive (\ie all gates are needed). \Ie if one input combination for one gate leads towards different model outputs, even though different gate outputs would be necessary, then the decision information is encoded into other masked inputs.
\end{compactenum}

\section{Results}
In this section we discuss our results, primarily for the DL models that reached 100\% accuracy on the split test set. Additional results on other settings can be found in Appendix~\ref{app:appendix}.

\subsection{General Performance and Tree Scores}
\begin{figure}[ht!]
	\centering
	\includegraphics[width=0.9999\columnwidth]{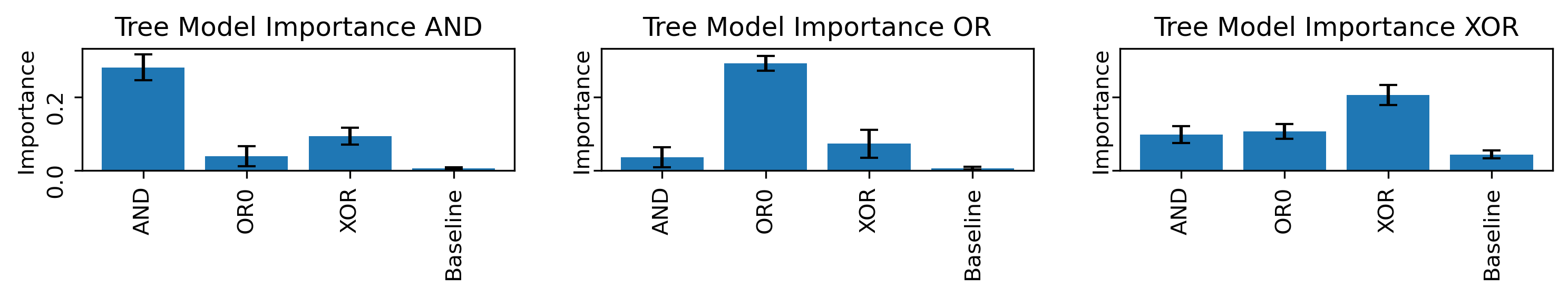}
	\caption{Average Random Forest import. with std. of the split test set, where the DL-Model reached 100\% acc.}
	\label{fig:treeImportance}
\end{figure}

On average the Random Forest performed slightly worse than the DL-Models, especially for the \textit{XOR}-top-level. Nevertheless, the tree-models output reasonable global relevancy distributions, as can be seen in Figure~\ref{fig:treeImportance}. Here, the \textit{Baseline} is very small compared to the rest. Notable is the dependency to the top-levels, where the gate equalling the top-level is always the most important; making sense logically because each lower gate has an easier time fulfilling the conditions of the top-level.

\subsection{Saliency Scores}
\begin{figure}[ht!]
	\centering
	\includegraphics[width=1.00\columnwidth]{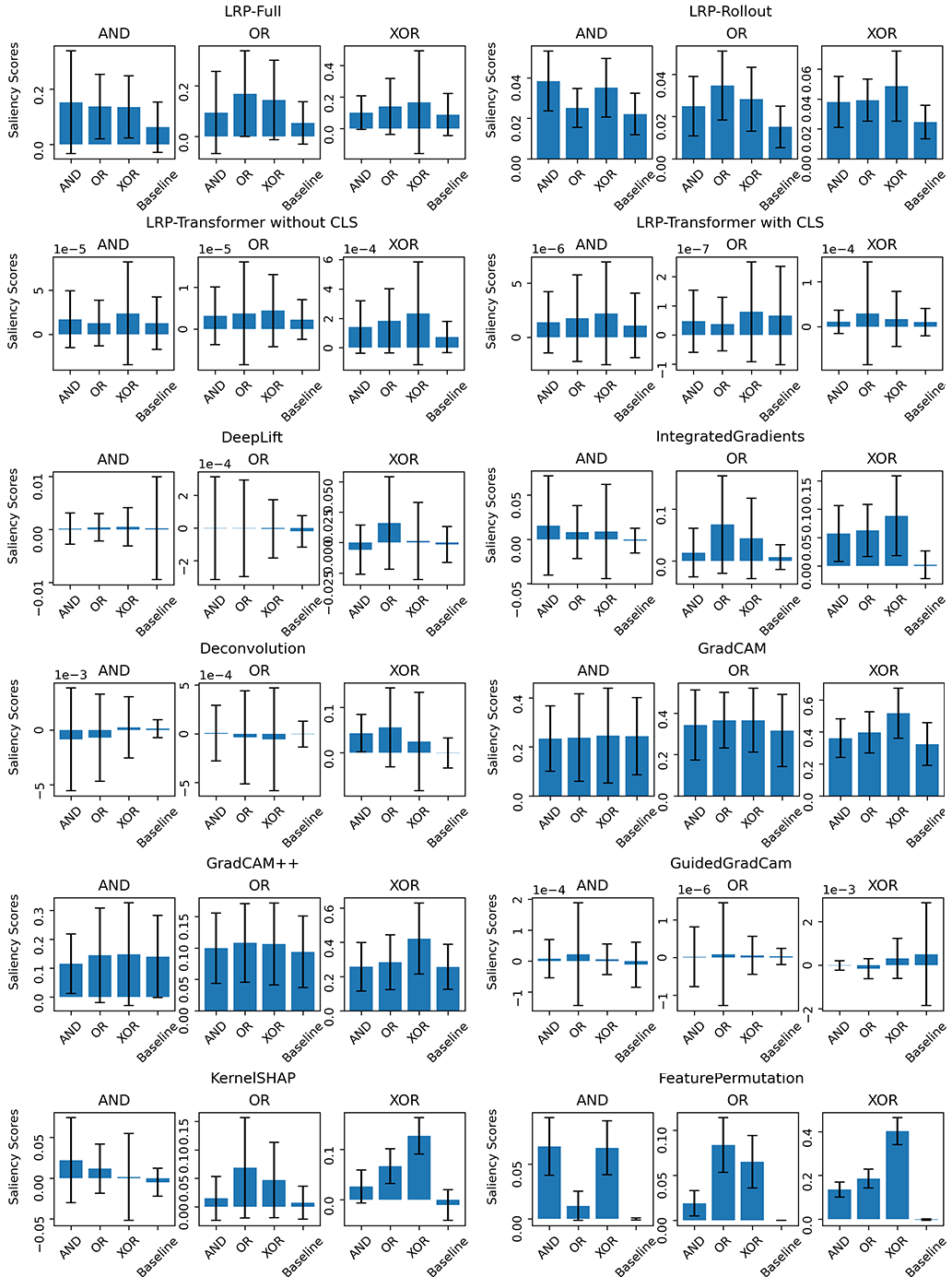}
	\caption{Average saliency scores with std. per logic gate per saliency method on the split test set between all trained DL-models, which reached a 100\% accuracy, \cf Figure \ref{fig:treeImportance}.}
	\label{fig:avgsaliencyScores}
\end{figure}

Compared to the tree scores, the average saliency scores per gate per method in Figure~\ref{fig:avgsaliencyScores} are less structured. The often quite highly scored \textit{Baseline}-block indicates Assumptions E and F are violated. Also notable is the high standard deviation that most methods and gates have (even when differentiating between classes -- see Appendix~\ref{app:saliencyResults}). This even occurs for the \textit{Baseline}-blocks, \ie a general consistency is missing, indicating either not globally comparable scores or an unknown undesired effect. The methods IntegratedGradients, FeaturePermutation and KernelSHAP are most similar to the tree importances, but only the FeaturePermutation has a consistent average saliency score ranking.

\begin{figure}[ht!]
	\centering
	\includegraphics[width=0.999\columnwidth]{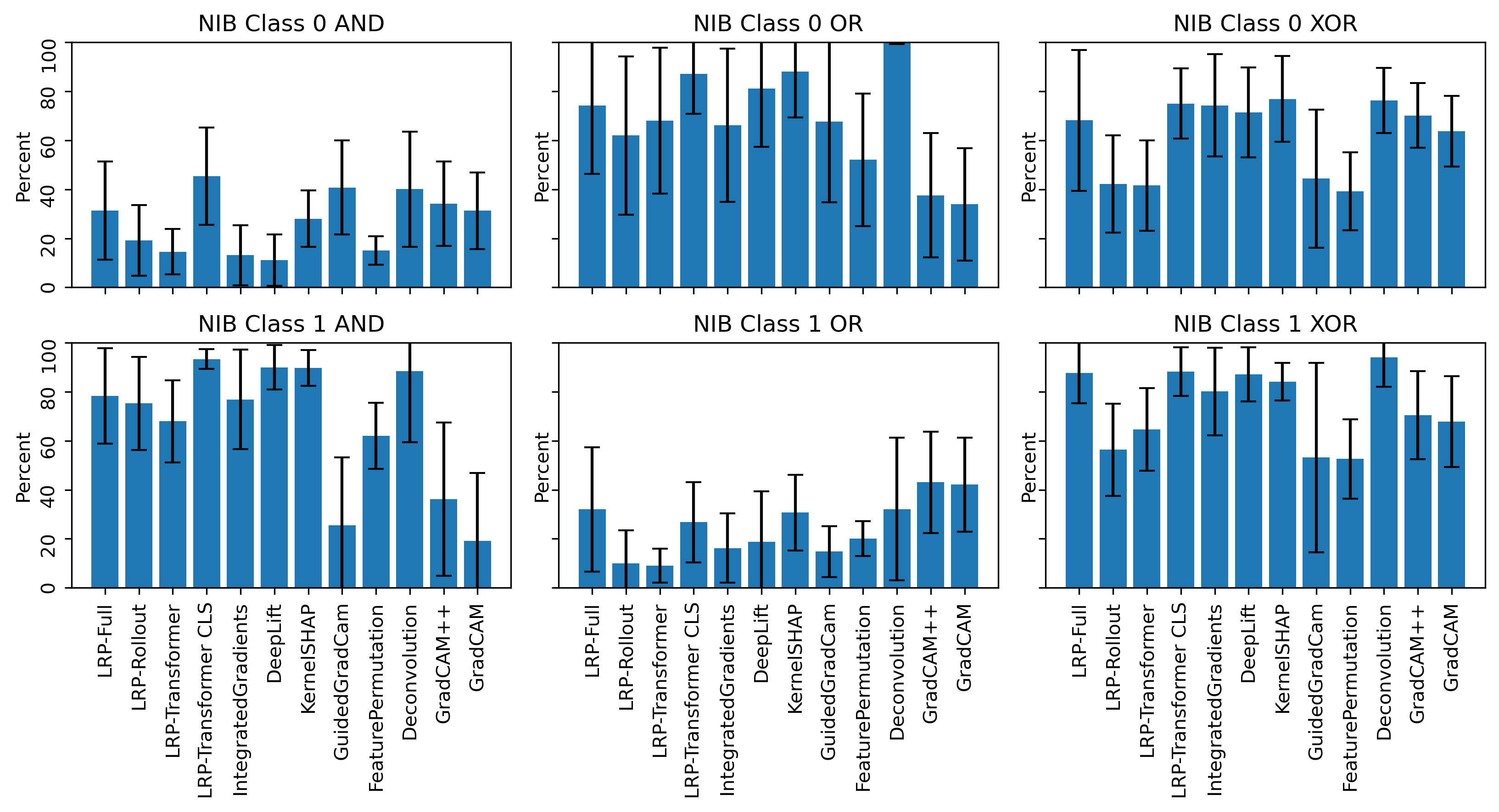}
	\caption{Average NIB with std. per class/top-level on the split test set of all DL-models with a 100\% accuracy.}
	\label{fig:avgNIB}
\end{figure}

The sample-wise evaluated NIB in Figure \ref{fig:avgNIB} shows many cases where the minimal information coverage is not maintained, \ie violating Assumptions E and F. This also means that
the general information coverage is also not given, \cf Appendix~\ref{app:retrainedResults}. Notable is the class influence on the \textit{OR}- and \textit{AND}-top-levels, \ie showing that information in the naturally more often occurring class is better approximated even though we balanced out the classes. This could be caused by a class bias favouring one class, \eg if it is not class $A$ so it is class $B$. However, the \textit{XOR}-top-level acts as possible counter-example.

\subsection{Retrained model results}

\begin{figure}[ht!]
	\centering
	\includegraphics[width=1.00\columnwidth]{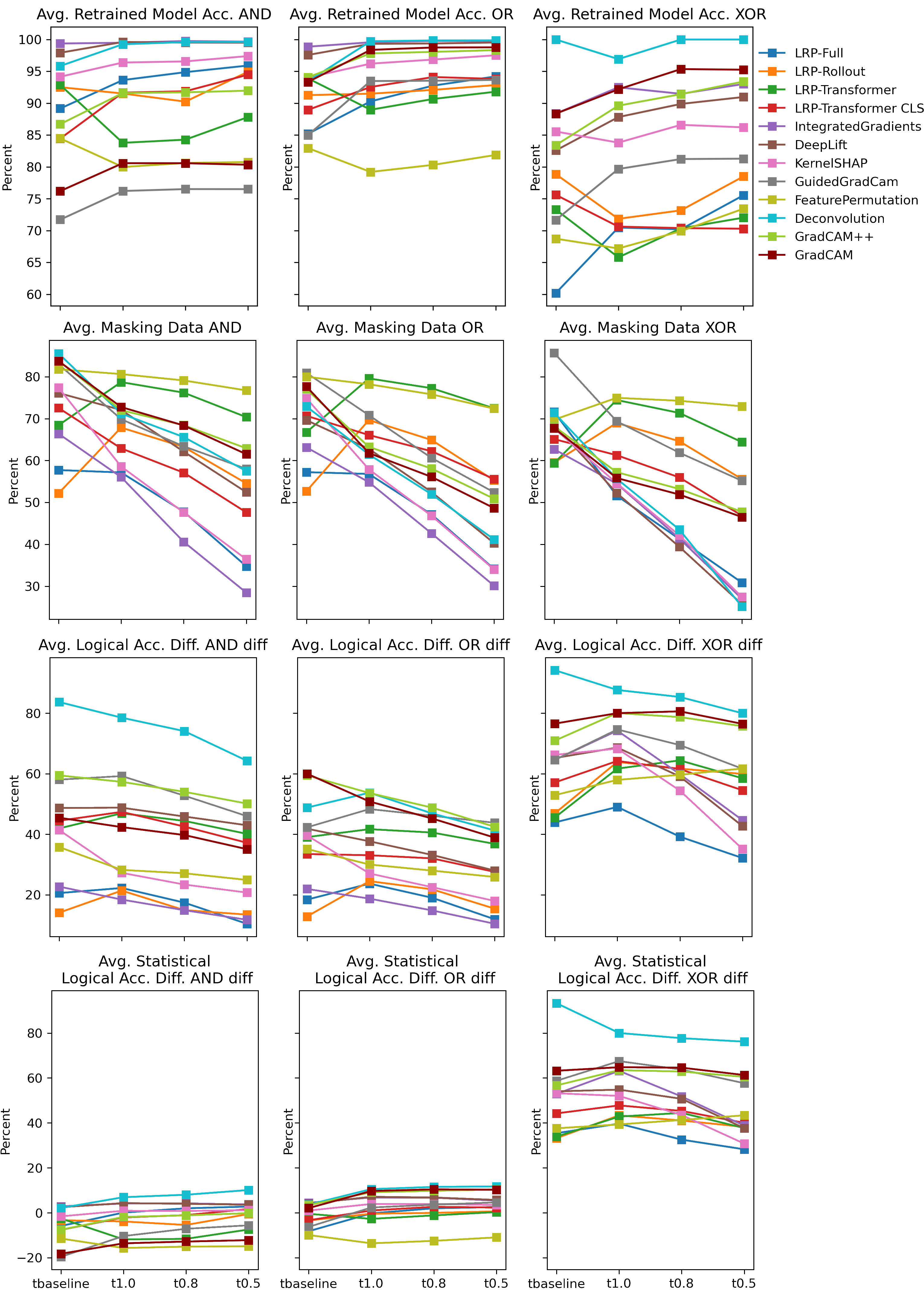}
	\caption{Avg. retrained model acc., masked data, logical acc. difference and statistical logical acc. difference (diff. to retrained model acc.) on the split test set for all DL-Models that reached 100\% acc.}
	\label{fig:accPerf}
\end{figure}

Figure \ref{fig:accPerf} shows the retrained model acc., the avg. percentage of masked inputs, as well as the difference in avg. acc. between the logical acc. and statistical logical acc, to the retrained acc. Notable is the masking and acc. ratio. Considering an ideal $R^d_{min}$ for each $x \in D$: the avg. reduction per test set can be at least 80\% for the \textit{AND}- and \textit{OR}-top-level and 40\% for the \textit{XOR}-top-level (for 2inBinary and 2inQuaternary),  without loosing any accuracy.  Additionally, the difference between the logical acc. (\ie the maintained information) and the retrained model is quite large. Hence, the methods do not mask optimally and this information must be included in the mask, \cf \citep{rong2022consistent}. 
This would be undesirable, since the real meaning of a sample would be hidden between complex combinations in the masked data w.r.t. to all inputs, while the inputs should be independent. The avg. statistical logical acc. diff. in Figure \ref{fig:accPerf} shows the possible to extract more classification information statistically, which reduces this difference around 0 (for \textit{AND} and \textit{OR}); while again the \textit{XOR}-top-level gives another example where a huge discrepancy exists. Additionally, the metric only shows the maximal reachable acc. and does not relate towards the model predictions. Here, the DCA results show that the models does actually capture different information.

\begin{figure}[ht!]
	\centering
	\includegraphics[width=0.70\columnwidth]{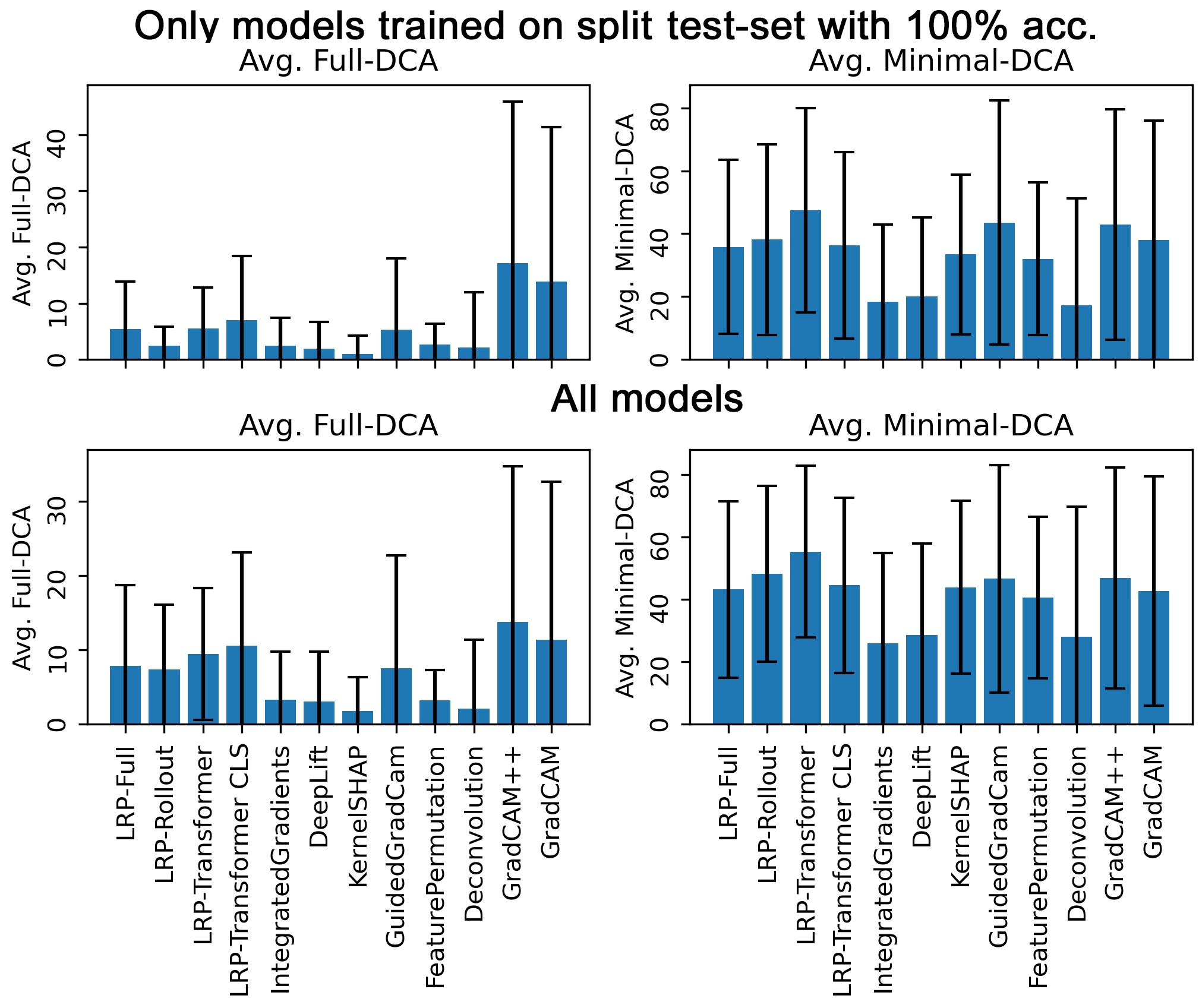}
	\caption{Avg DCAs with std. of the split test set with only DL-models reaching 100\% acc. (top), compared to DCAs of all trained models reaching 100\% acc. (bottom), compared to DCAs of all trained), showing the similarity between those conditions.}
	\label{fig:avgdca}
\end{figure}

\begin{figure*}[ht!]
	\centering
	\includegraphics[width=1.42\columnwidth]{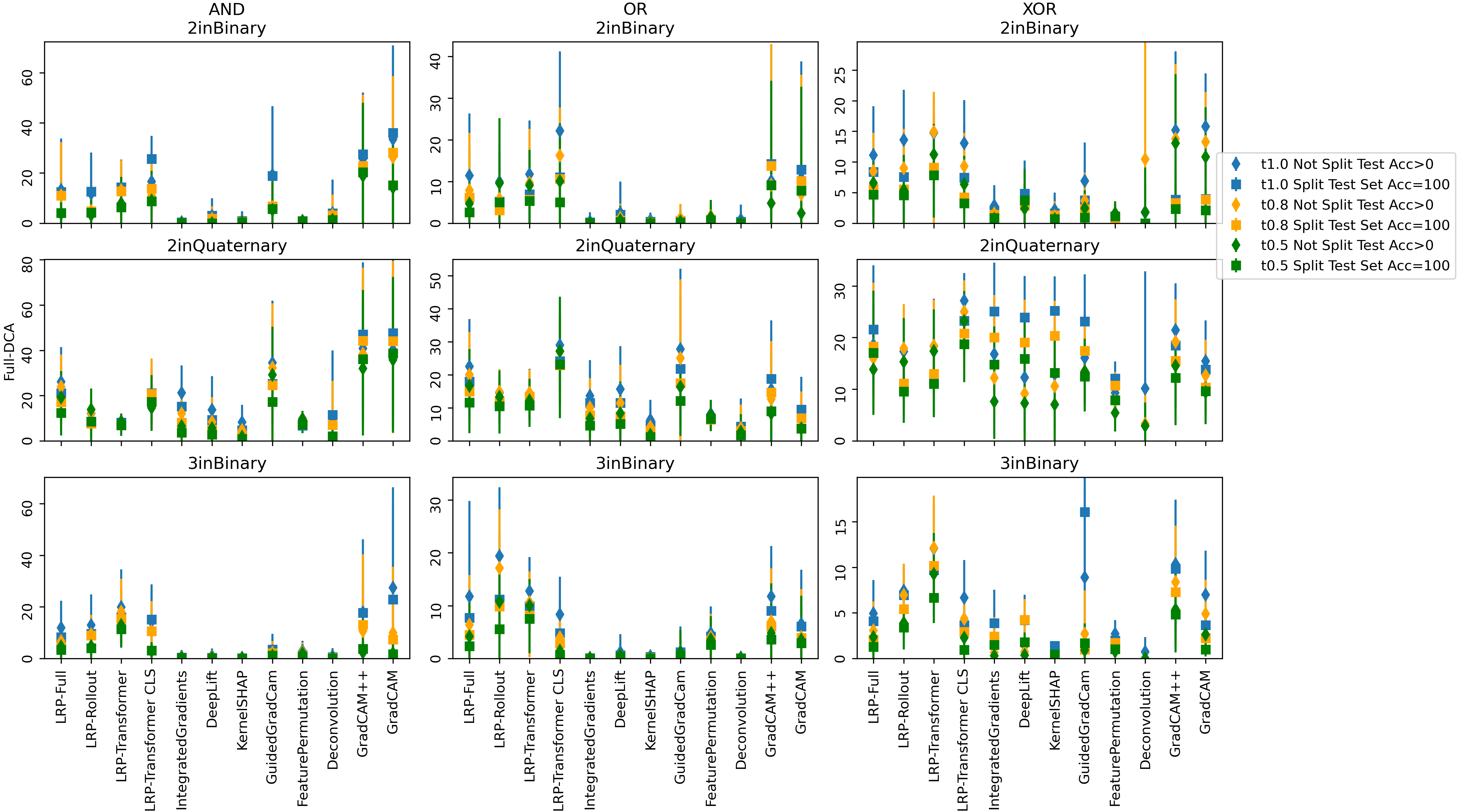}
	\caption{Full DCA with std., showing how often relevant combinations occur as positive and as negative class after masking, based on 4 different thresholds per saliency method; done on all (Split) test data samples for base models that reached 100\% acc and on all models where the train- and test set contain all samples (Not Split).}
	\label{fig:dcaFULL}
\end{figure*}

This encoding problem is further emphasized in Figure~\ref{fig:dcaFULL}, showing the average Full-DCA. While for simple datasets some methods (IntegratedGradients, DeepLift, KernelSHAP, FeaturePermutation, Deconvolution) rarely encode information into the Baseline, for the \textit{XOR}-top-level and especially the 2inQuaternary dataset (most combinations), the mask consistently encodes information into the \textit{Baseline} inputs. \citet{shah2021input} called some similar phenomena, information leakage, but we argue that this is some form of internal model encoding. A higher threshold seems to increase this leakage. This makes sense, but considering, that the reduction could be higher, this just further highlights that each method does not capture information as expected and encodes information in lower scores. 
Figures \ref{fig:FullDCAAll} and \ref{fig:MinDCAAll} in the Appendix show similar DCA ranges (but somewhat higher, but probably due to the bigger sample size) over the experiments that do not reach a 100\% test acc. and where the train and test set contain all samples. Figure \ref{fig:avgdca} further strengthens this assumption, by showing  consistent and comparable differences between the average DCAs over all DL-models with 100\% acc. on the split test set (top), to the overall average over all trained models (bottom). Further, Figure \ref{fig:avgdca} shows that the minimal DCA is often quite high, \ie relevant information is often distributed towards the locally irrelevant gate inputs. For IntegratedGradients, DeepLift and Deconvolution this effect occurred less often, but considering the full data (Appendix Figure \ref{fig:MinDCAAll}), this effect is again very consistent on the 2inQuaternary dataset. Consequentially, this encoding effect could always potentially occur, and thus limit the trustworthiness of those saliency methods in general. Overall, IntegratedGradients had the best logical acc. diff. to DCA ratio. However, the DCA only considers specific cases, \eg Deconvolution has low DCAs but a very high logical acc. diff., \ie multiple other forms of information encoding can exist, which are harder to test for.

\section{Discussion and Limitations}
Although all methods failed to capture the NIB, each retrained model showed better performance than expected. Considering that relevant input combinations can lead towards different model outputs (\cf DCA), this can be attributed to a suboptimal ranking order of saliency scores, introducing an encoding more complex than just \eg image outlines \citep{rong2022consistent}. Consequentially, the extracted relevancy cannot be interpreted just by the numeric value scale, in contrast to our assumptions. We tested this over multiple thresholds, datasets and conditions, which showed consistent occurrences and ranges per method for this encoding.
The reason for this encoding is, however, still unclear, if it is a bias in the methods, due to internal procedure of the model --- \eg internal numeric potentials --- or if too much information is aggregated (\cf \citep{harris2021joint, kumar2021shapley}). While methods like SHAP and FeaturePermutation are model agnostic and mathematically well-founded, they still run into problems~\citep{slack2020fooling, fryer2021shapley, ju2021logic, hooker2021unrestricted, harris2021joint, kumar2021shapley}.
Nonetheless, saliency methods already proved they can be useful for certain scenarios \eg for bias detection~\citep{lapuschkin2019unmasking}. This opens up the question, in which conditions saliency scores are fully trustworthy.

It remains unclear, if it is possible to reliably decode all relevancy information. Maybe it is not possible to work with masks at all (\cf \citep{sturmfels2020visualizing}), because inputs cannot be completely ignored~\citep{rong2022consistent} and hence more sequential-based evaluation (the ability to ignore inputs), logical based explanations or more complex methods aggregating less information are needed.
While we found certain methods encode data less often, they might perform differently on different tasks~\citep{kokhlikyan2021investigating, yona2021revisiting}. Nonetheless, we argue, that due to the basic relation contain \andor, similar cases are contained in complex realistic datasets.

\section{Conclusion and Future Work}
In this paper, we introduced a propositional logic-based dataset framework \andor that employs a local saliency baseline and can approximate ground truth reasoning. It can be used to analyse different scenarios of information relations (complementary, redundant, exclusive), as well as to verify the local minimal information coverage of saliency methods. Furthermore, we presented several metrics for evaluation.

Our experiments show that a global aggregation of the local saliency scores are often inconsistent, and only three methods could somewhat find plausible global relations. Further, our results indicate that all methods fail to capture all the relevant local information (\cf NIB) and even encode class discriminative information into the saliency score ranking order (\cf DCA). While this effect is less prominent for certain methods on simpler tasks, especially on our most complex task (2inQuaternary XOR-top-level) this effect is very consistent for all
methods; consequentially limiting the trust of saliency methods for even more complex tasks.

We argue that our testing framework can be applied as a trust and quality test for different information relation scenarios -- considering saliency-based methods. This is important, in particular, for the verification and improvement relating to explainable AI and explainable and interpretable machine learning methods.
For future work, we aim to analyse the reasons for encoding, perform more experiments on singular logical operations, as well as perform more studies towards logical and sequential explanations.

\newpage

\bibliographystyle{named}
\bibliography{main}

\clearpage
\appendix
\FloatBarrier

\onecolumn
\section{Appendix}
\label{app:appendix}
In the following, we provide addition plots, mostly for all trained models, to provide further results supporting our discussion.

\subsection{General Performance}
\label{app:generalPerformance}
\FloatBarrier
\begin{figure}[htb!]
	\centering
	\includegraphics[width=.74\columnwidth]{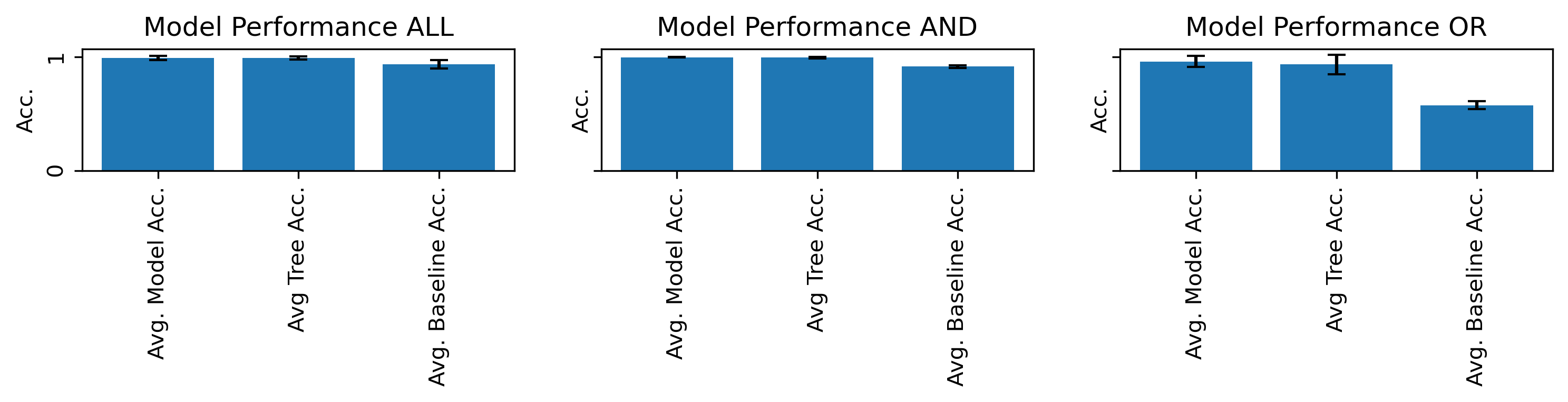}
	\caption{Average Acc. over all trained models, including a baseline.}
	\label{fig:generallAcc}
\end{figure}

\begin{figure}[htb!]
	\centering
	\includegraphics[width=.74\columnwidth]{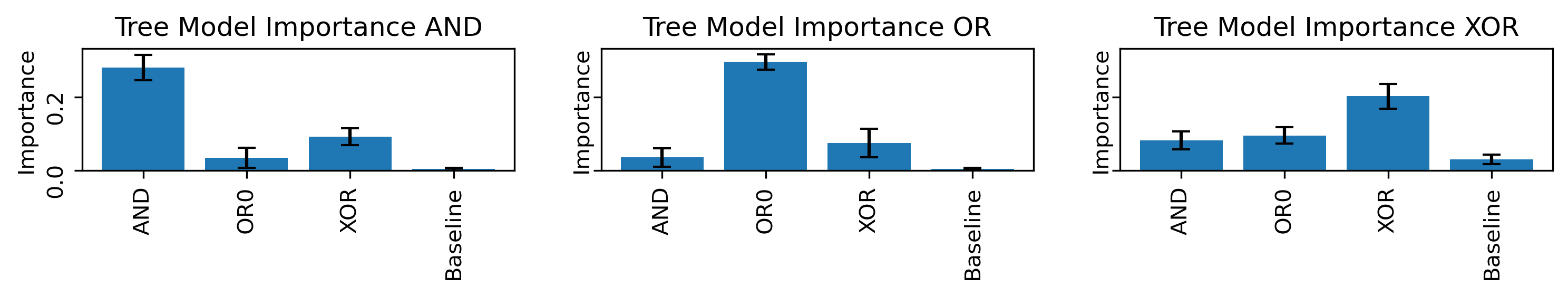}
	\caption{Average tree importance over all trained models.}
	\label{fig:generallTreeImp}
\end{figure}

\FloatBarrier
\subsection{Saliency Results}
\label{app:saliencyResults}
\FloatBarrier
\begin{figure*}[htb!]
	\centering
	\includegraphics[width=.93\columnwidth]{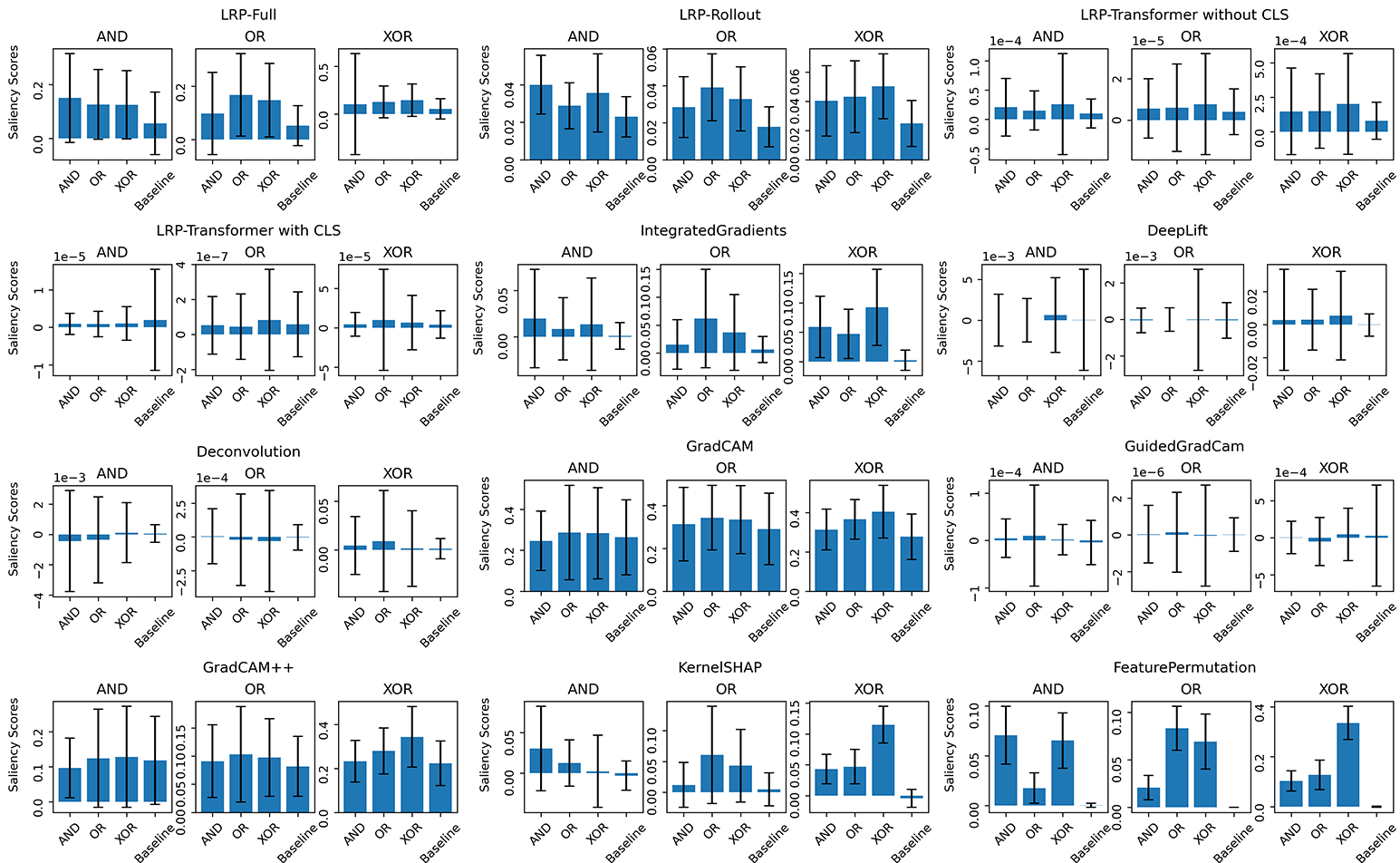}
	\caption{Average saliency scores per logic gate per saliency method, based on all trained DL-models.}
	\label{fig:avgsaliencyAll}
\end{figure*}

\FloatBarrier

\begin{figure}[htb!]
	\centering
	\includegraphics[width=.95\columnwidth]{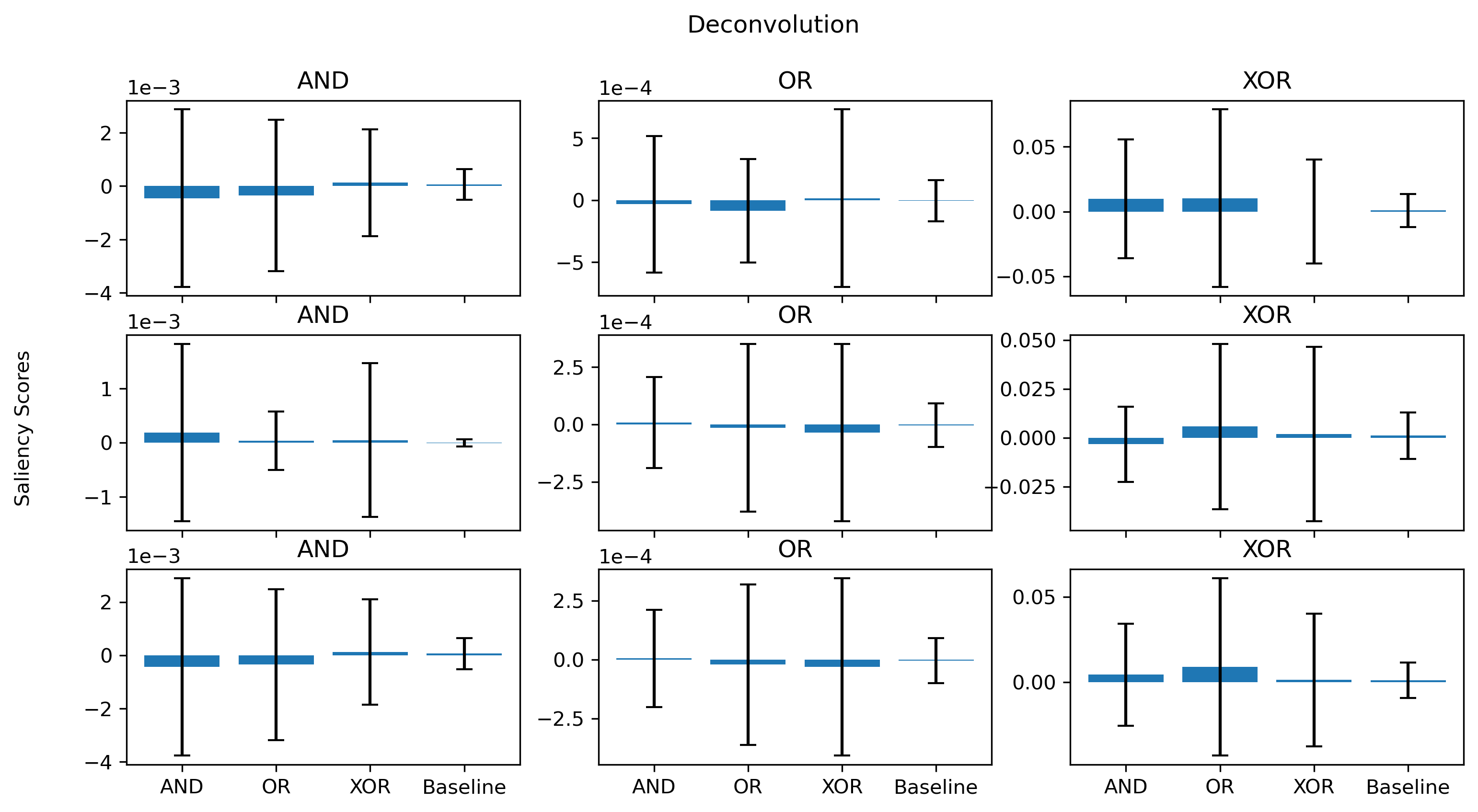}
	\caption{Average saliency scores per class per logic gate for Deconvolution, based on all trained DL-models. Classes are separated as the following: class 0 (top row), class 1 (middle row) and overall average (bottom row).}
	\label{fig:avgsaliencyDeconv}
\end{figure}

\begin{figure}[htb!]
	\centering
	\includegraphics[width=.95\columnwidth]{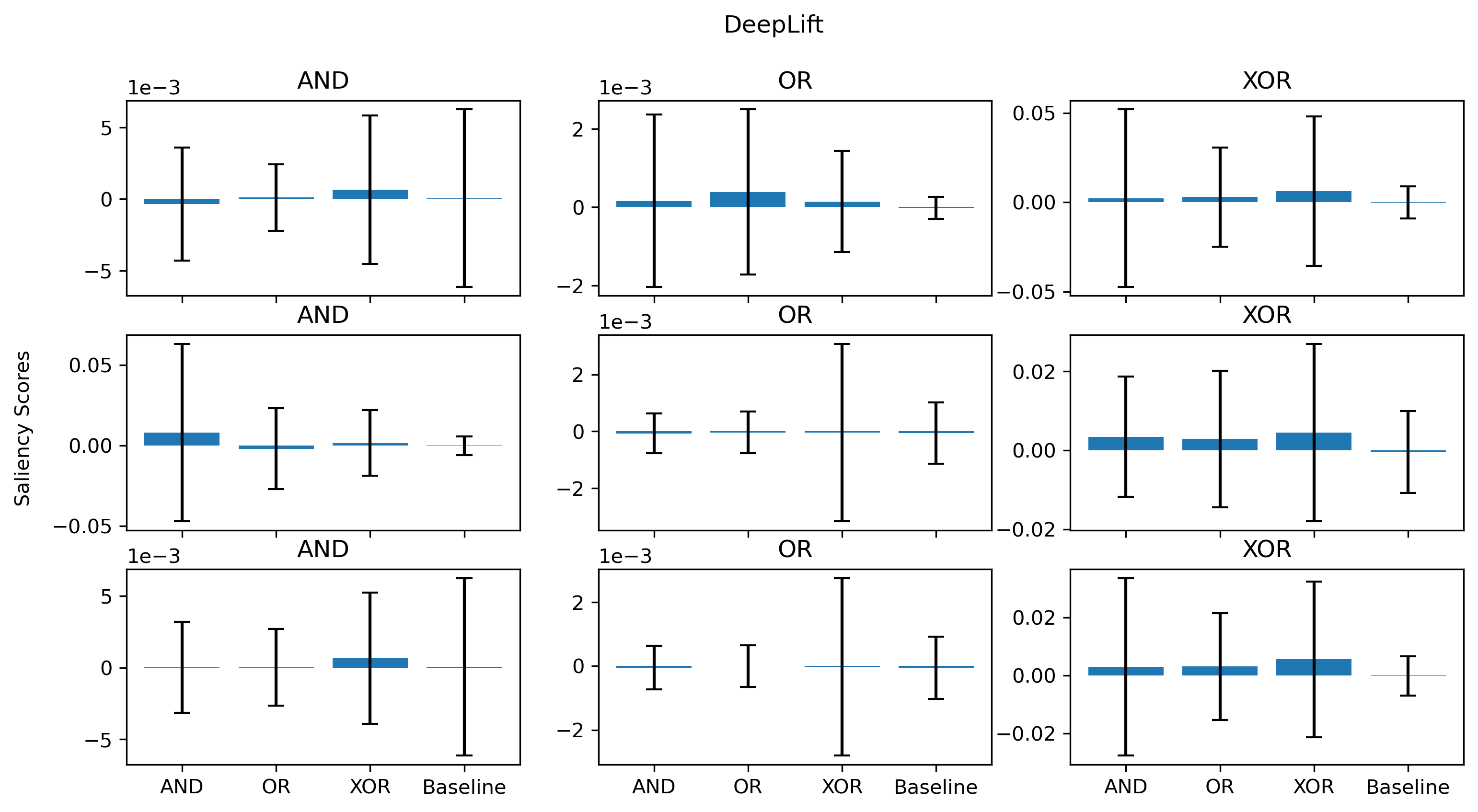}
	\caption{Average saliency scores per class per logic gate for DeepLift, based on all trained DL-models. Classes are separated as the following: class 0 (top row), class 1 (middle row) and overall average (bottom row).}
	\label{fig:avgsaliencyDeepLift}
\end{figure}

\begin{figure}[htb!]
	\centering
	\includegraphics[width=.95\columnwidth]{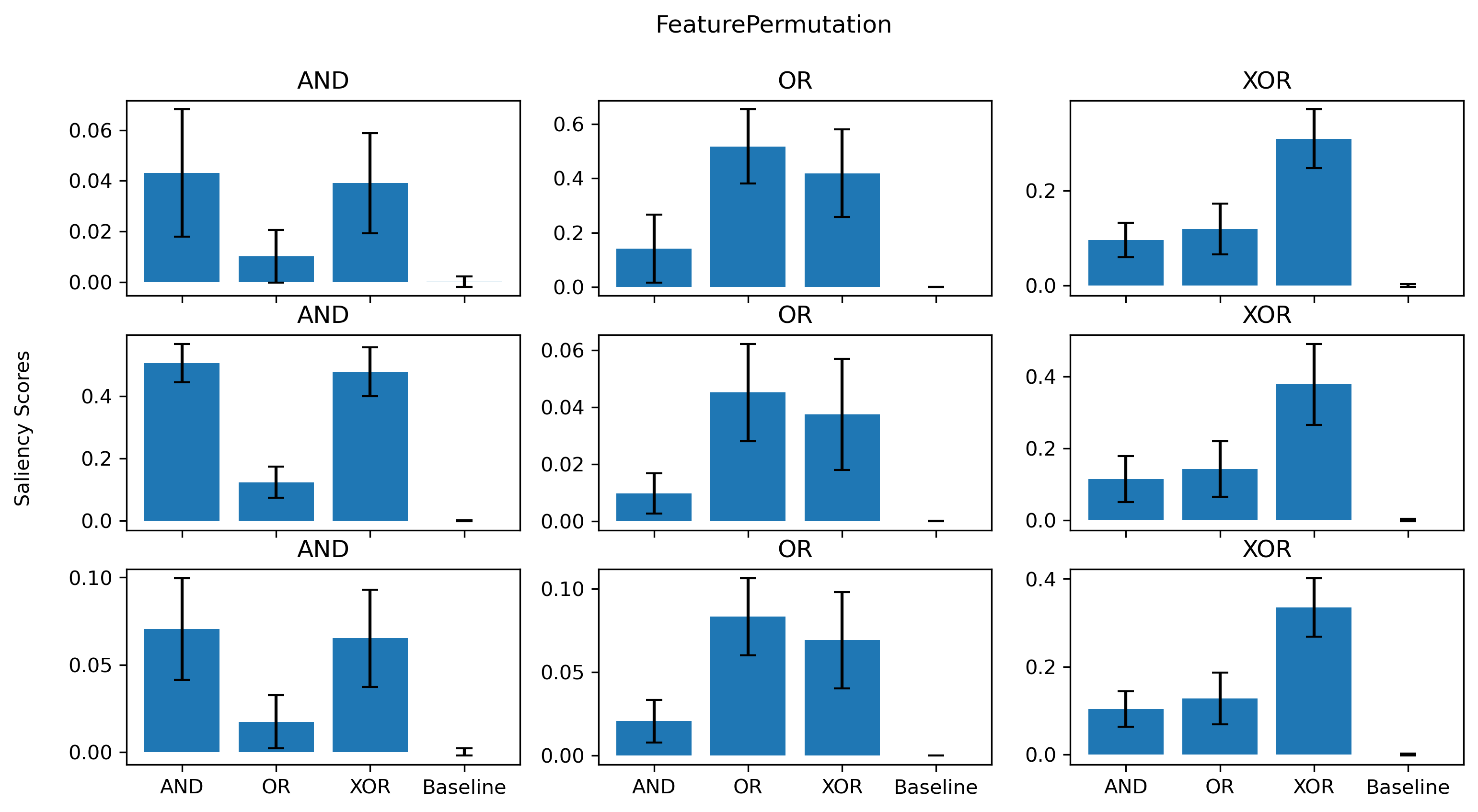}
	\caption{Average saliency scores per class per logic gate for FeaturePermutation, based on all trained DL-models. Classes are separated as the following: class 0 (top row), class 1 (middle row) and overall average (bottom row).}
	\label{fig:avgsaliencyFP}
\end{figure}

\begin{figure}[htb!]
	\centering
	\includegraphics[width=.95\columnwidth]{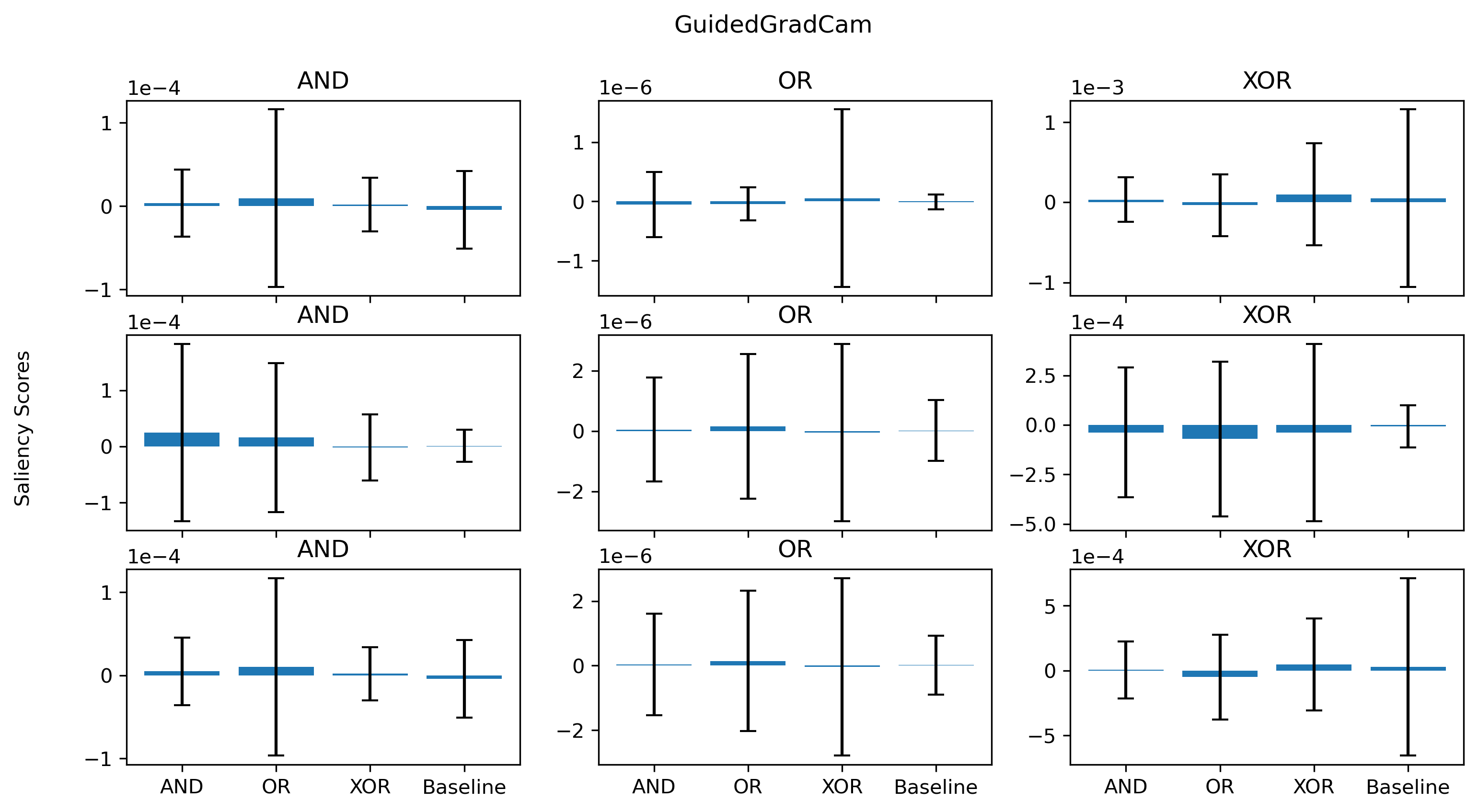}
	\caption{Average saliency scores per class per logic gate for GuidedGradCam, based on all trained DL-models. Classes are separated as the following: class 0 (top row), class 1 (middle row) and overall average (bottom row).}
	\label{fig:avgsaliencyGuidedGradCam}
\end{figure}

\begin{figure}[htb!]
	\centering
	\includegraphics[width=.95\columnwidth]{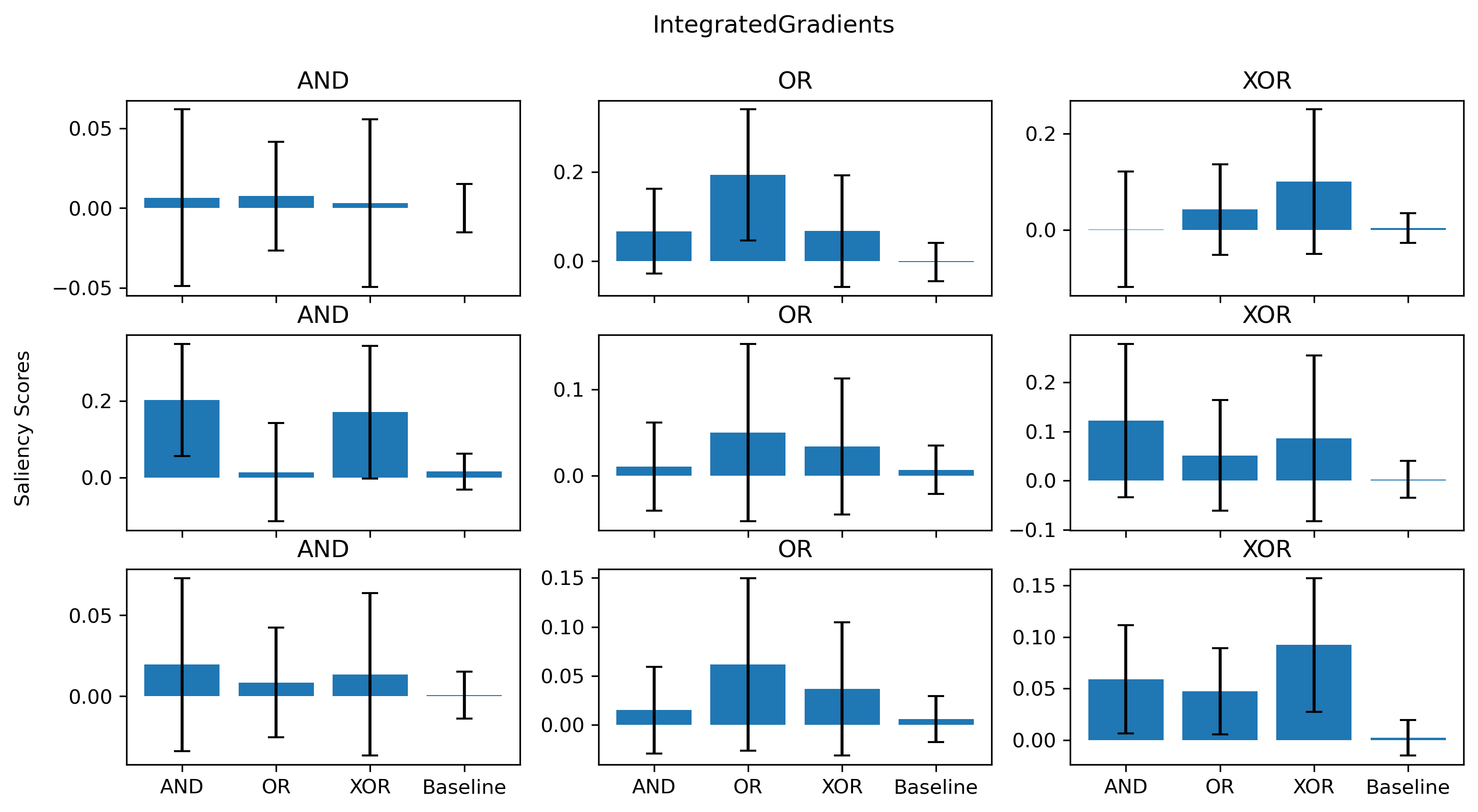}
	\caption{Average saliency scores per class per logic gate for IntegratedGradients, based on all trained DL-models. Classes are separated as the following: class 0 (top row), class 1 (middle row) and overall average (bottom row).}
	\label{fig:avgsaliencyIG}
\end{figure}

\begin{figure}[htb!]
	\centering
	\includegraphics[width=.95\columnwidth]{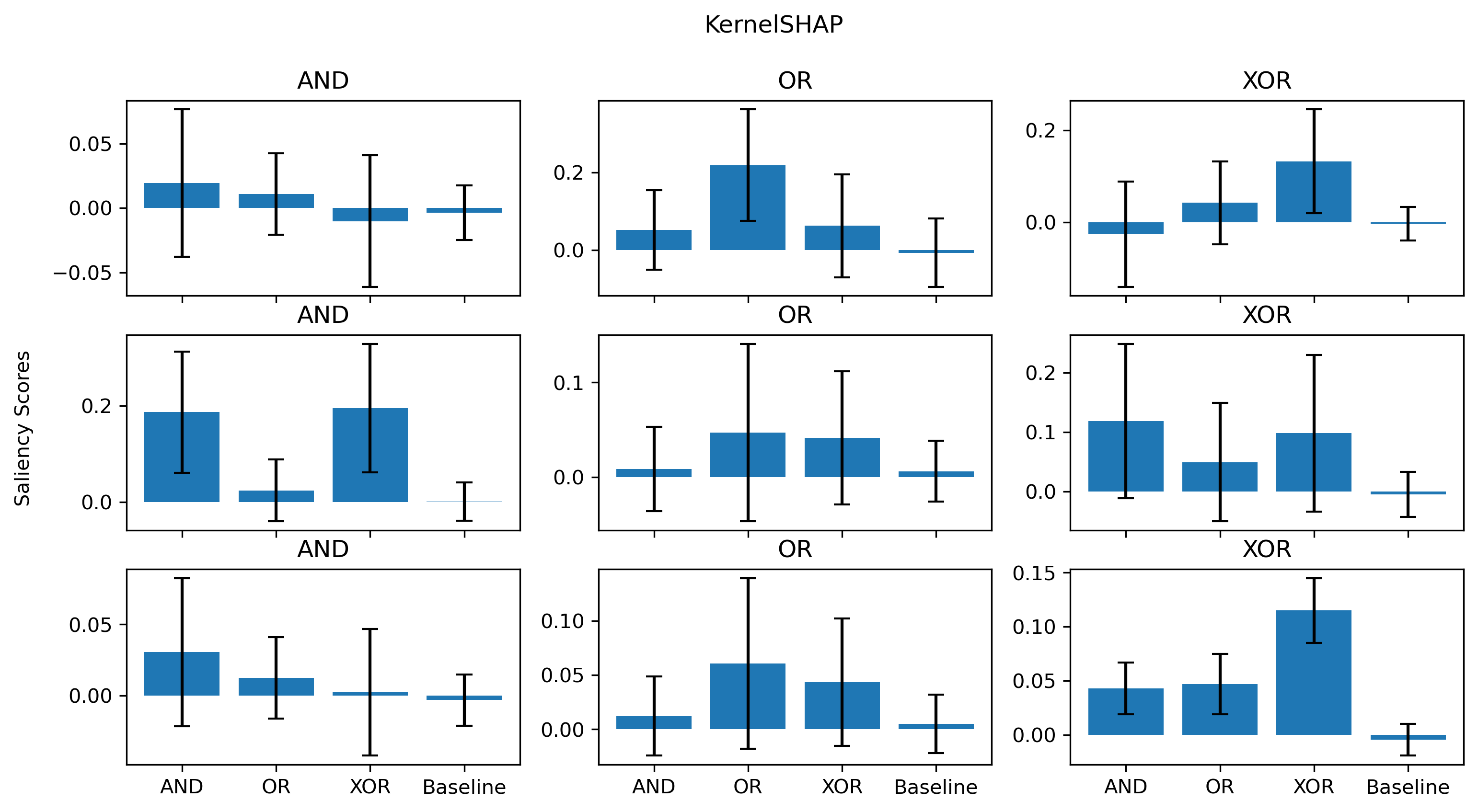}
	\caption{Average saliency scores per class per logic gate for KernalSHAP, based on all trained DL-models. Classes are separated as the following: class 0 (top row), class 1 (middle row) and overall average (bottom row).}
	\label{fig:avgsaliencyShap}
\end{figure}

\begin{figure}[htb!]
	\centering
	\includegraphics[width=.95\columnwidth]{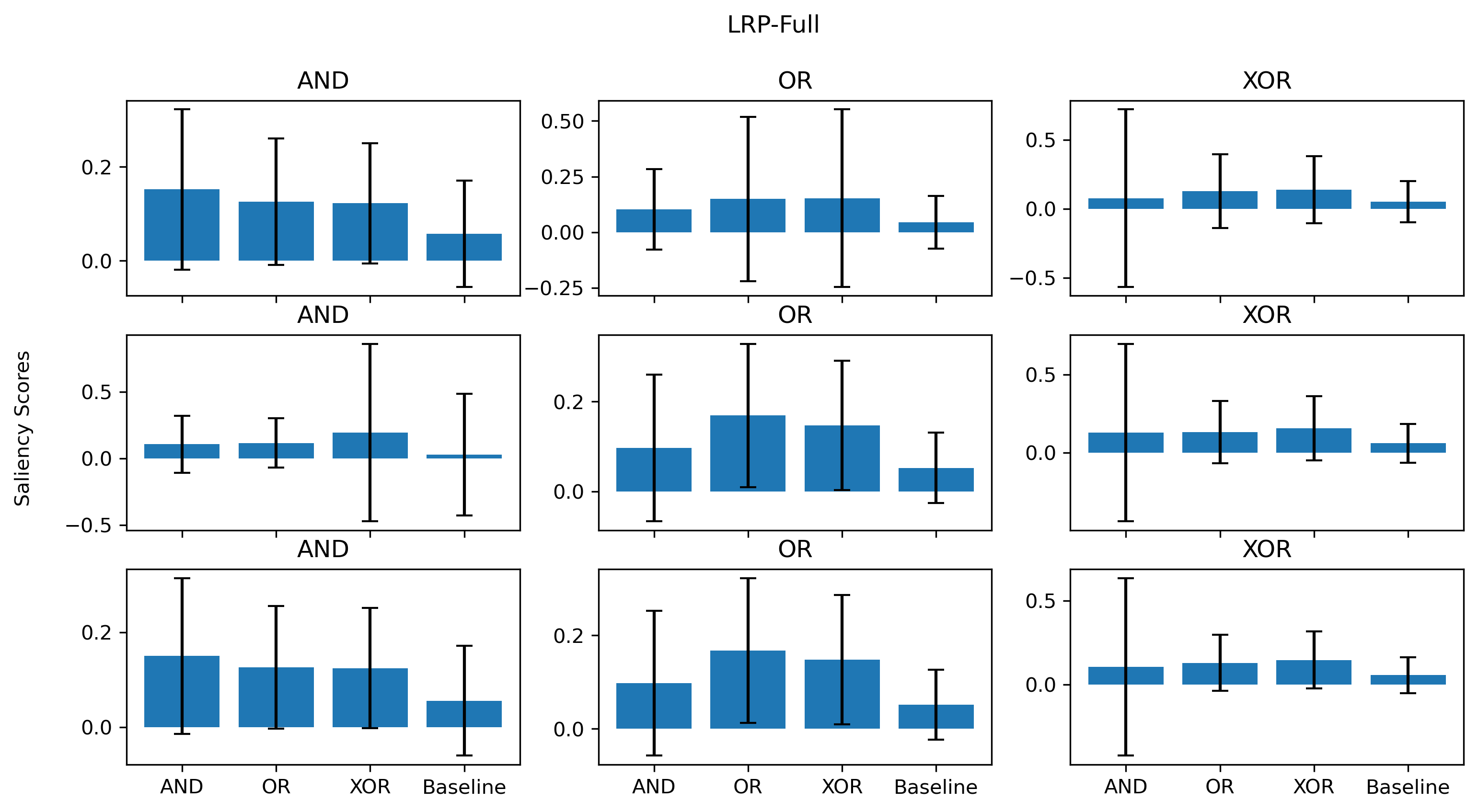}
	\caption{Average saliency scores per class per logic gate for LRP-Full, based on all trained DL-models. Classes are separated as the following: class 0 (top row), class 1 (middle row) and overall average (bottom row).}
	\label{fig:avgsaliencyLRPFull}
\end{figure}

\begin{figure}[htb!]
	\centering
	\includegraphics[width=.95\columnwidth]{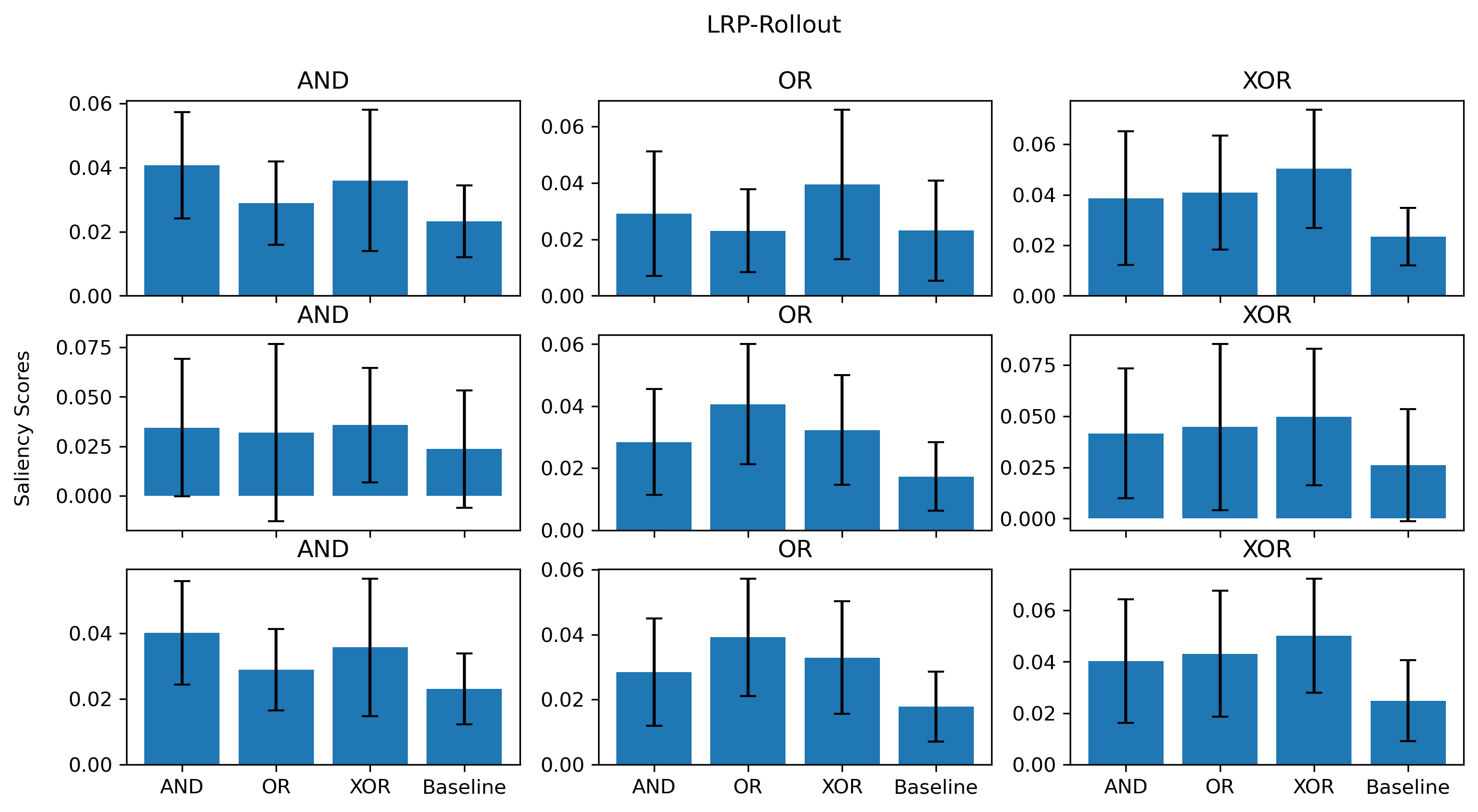}
	\caption{Average saliency scores per class per logic gate for LRP-Rollout, based on all trained DL-models. Classes are separated as the following: class 0 (top row), class 1 (middle row) and overall average (bottom row).}
	\label{fig:avgsaliencyLRPRollout}
\end{figure}

\begin{figure}[htb!]
	\centering
	\includegraphics[width=.95\columnwidth]{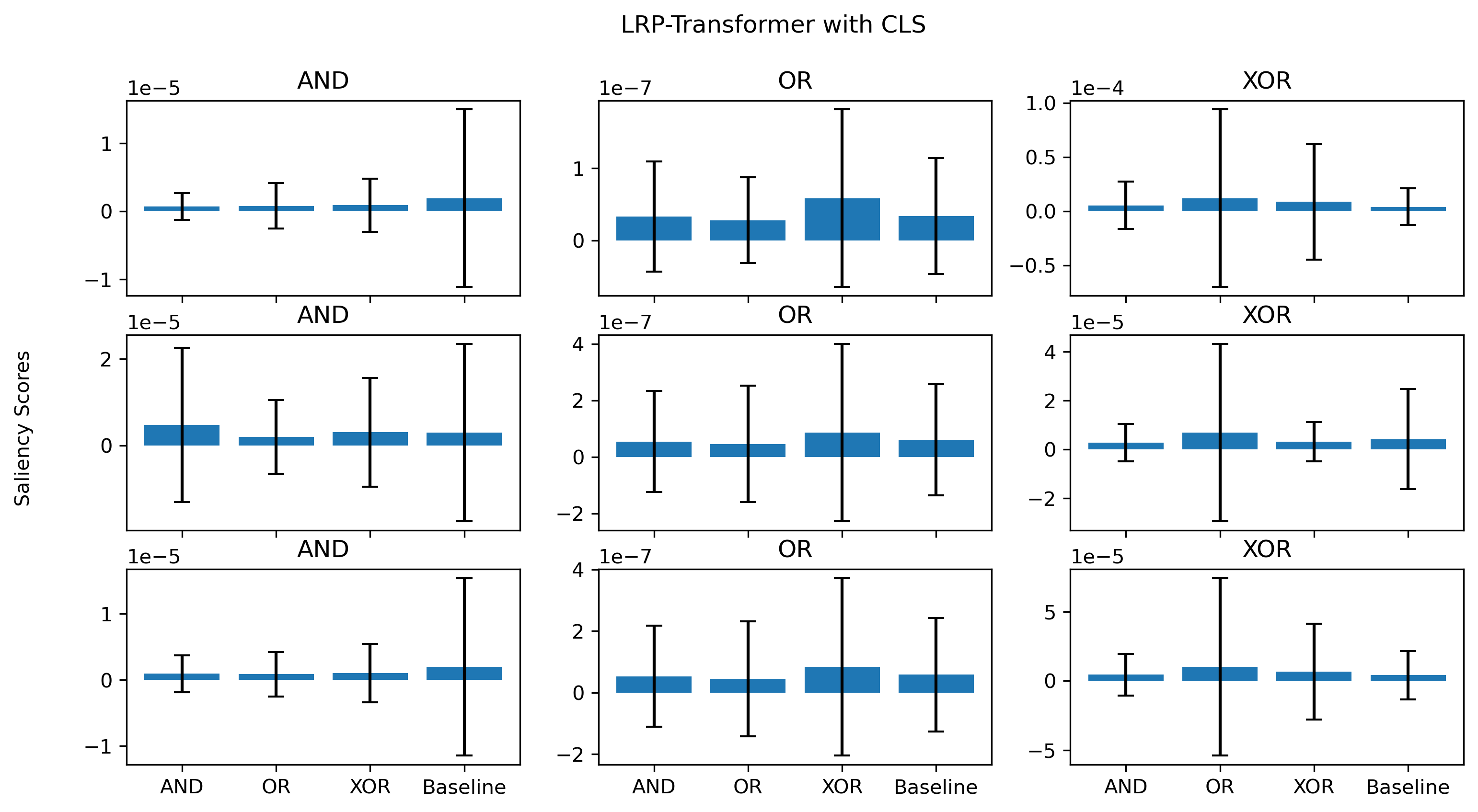}
	\caption{Average saliency scores per class per logic gate for Transformer attribution with CLS, based on all trained DL-models. Classes are separated as the following: class 0 (top row), class 1 (middle row) and overall average (bottom row).}
	\label{fig:avgsaliencyLRPAttCls}
\end{figure}

\begin{figure}[hbt!]
	\centering
	\includegraphics[width=.95\columnwidth]{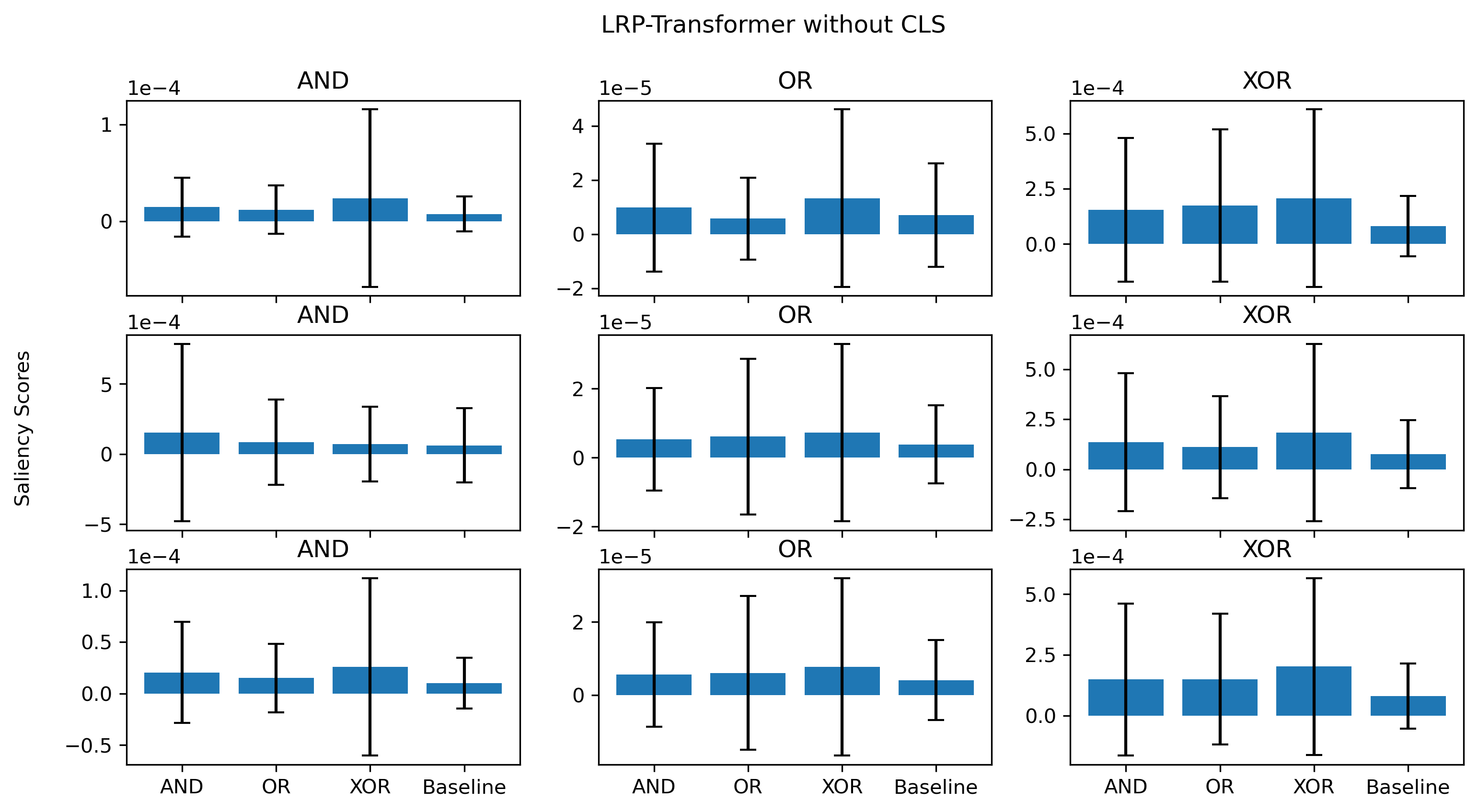}
	\caption{Average saliency scores per class per logic gate for Transformer attribution without CLS, based on all trained DL-models. Classes are separated as the following: class 0 (top row), class 1 (middle row) and overall average (bottom row).}
	\label{fig:avgsaliencyLRPAttNoCls}
\end{figure}

\begin{figure}[htb!]
	\centering
	\includegraphics[width=.95\columnwidth]{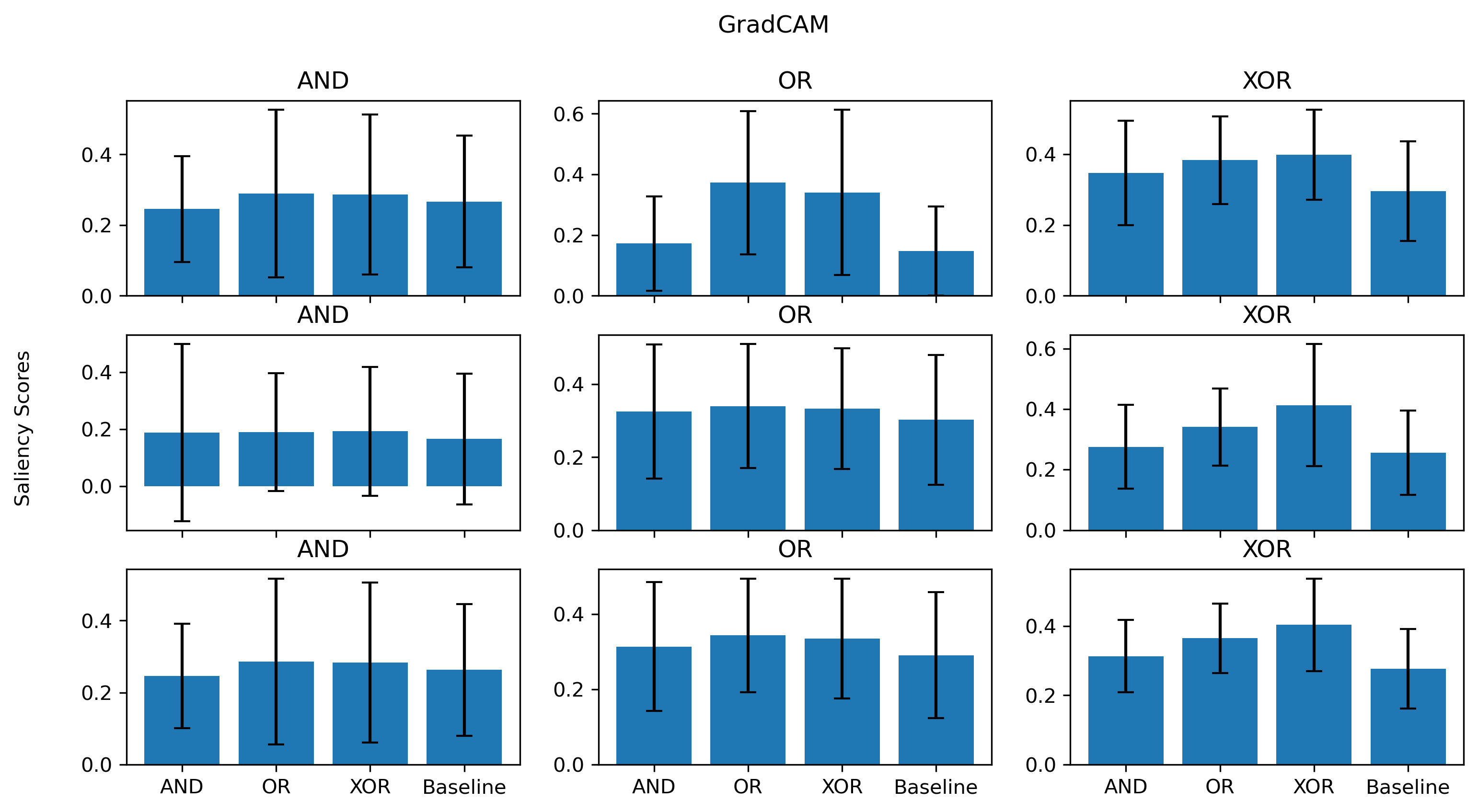}
	\caption{Average saliency scores per class per logic gate for GradCAM, based on all trained DL-models. Classes are separated as the following: class 0 (top row), class 1 (middle row) and overall average (bottom row).}
	\label{fig:avgsaliencyGradCAM}
\end{figure}

\begin{figure}[htb!]
	\centering
	\includegraphics[width=.95\columnwidth]{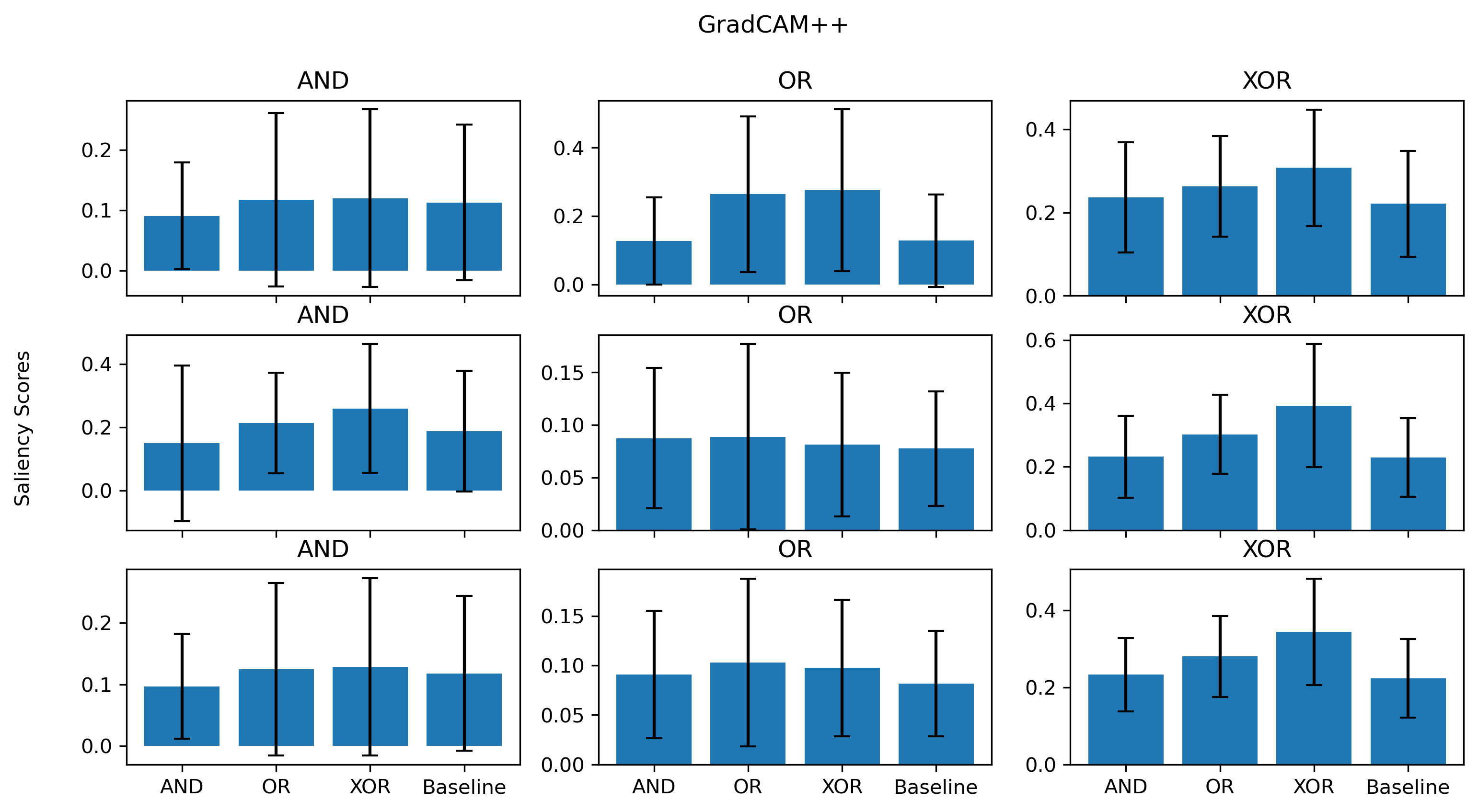}
	\caption{Average saliency scores per class per logic gate for GradCAM++, based on all trained DL-models. Classes are separated as the following: class 0 (top row), class 1 (middle row) and overall average (bottom row).}
	\label{fig:avgsaliencyGradCAMpp}
\end{figure}

\begin{figure}[htb!]
	\centering
	\includegraphics[width=.99\columnwidth]{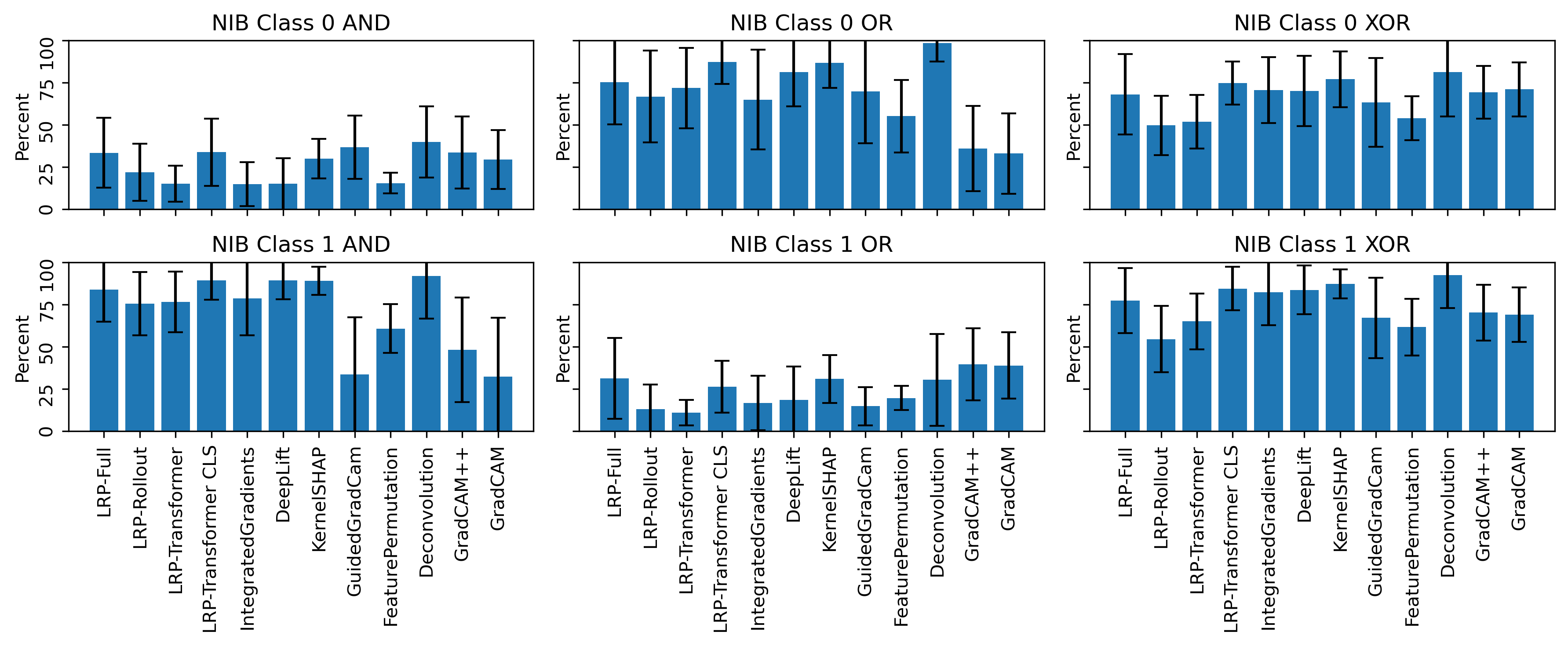}
	\caption{Average NIB per class on test sets of all trained DL-models.}
	\label{fig:NIBAll}
\end{figure}

\begin{figure}[htb!]
	\centering
	\includegraphics[width=.99\columnwidth]{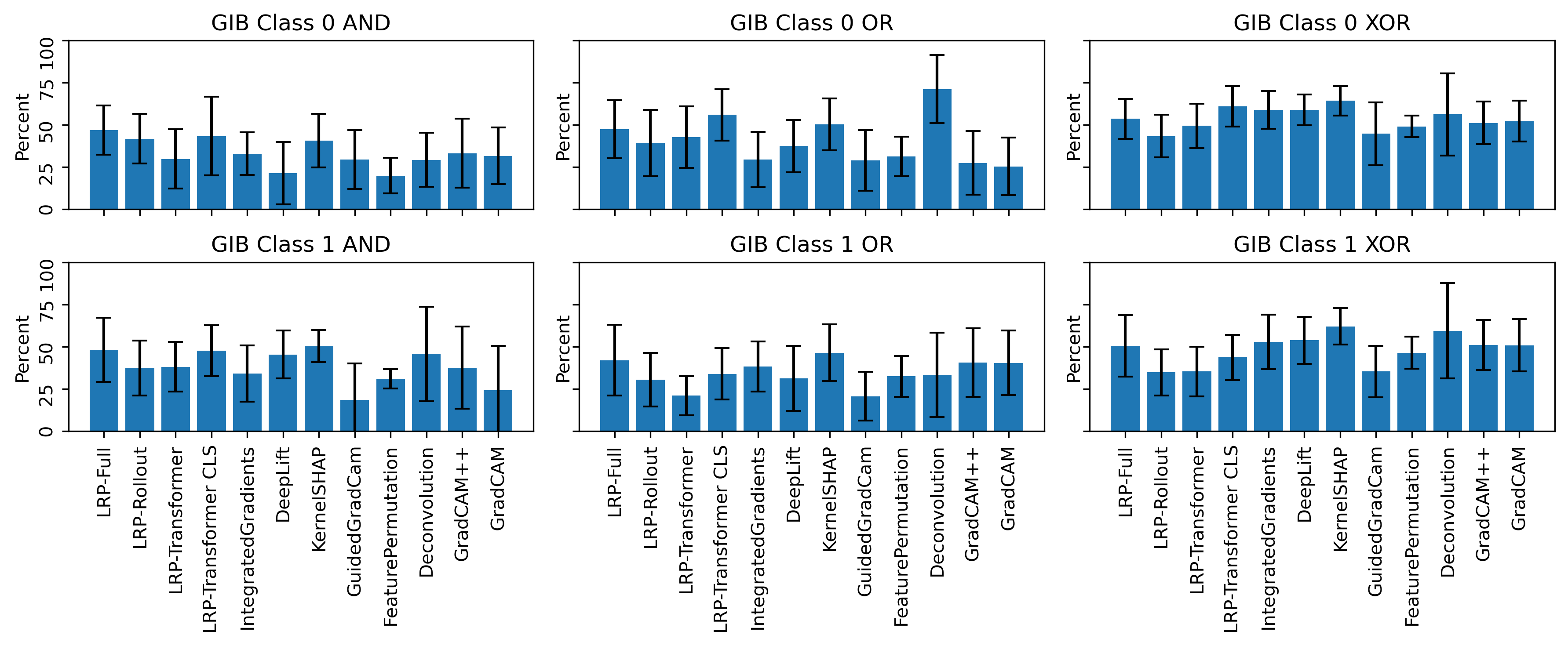}
	\caption{Average GIB per class on test sets of all trained DL-models.}
	\label{fig:GIBAll}
\end{figure}

\begin{figure}[htb!]
	\centering
	\includegraphics[width=.99\columnwidth]{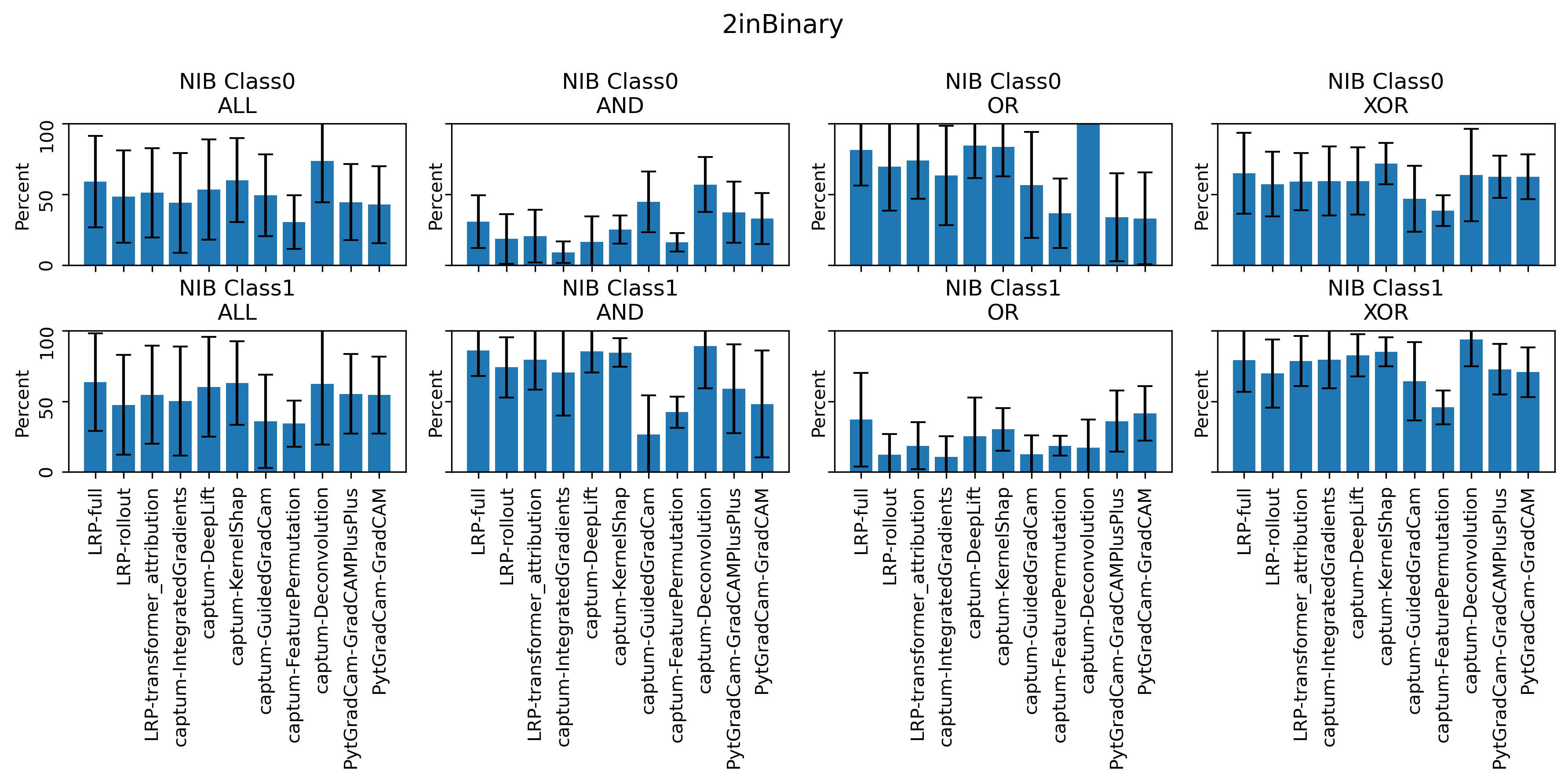}
	\caption{Average NIB per class on test sets of all trained DL-models on the 2inBinary dataset.}
	\label{fig:NIB2inBinary}
\end{figure}

\begin{figure}[htb!]
	\centering
	\includegraphics[width=.99\columnwidth]{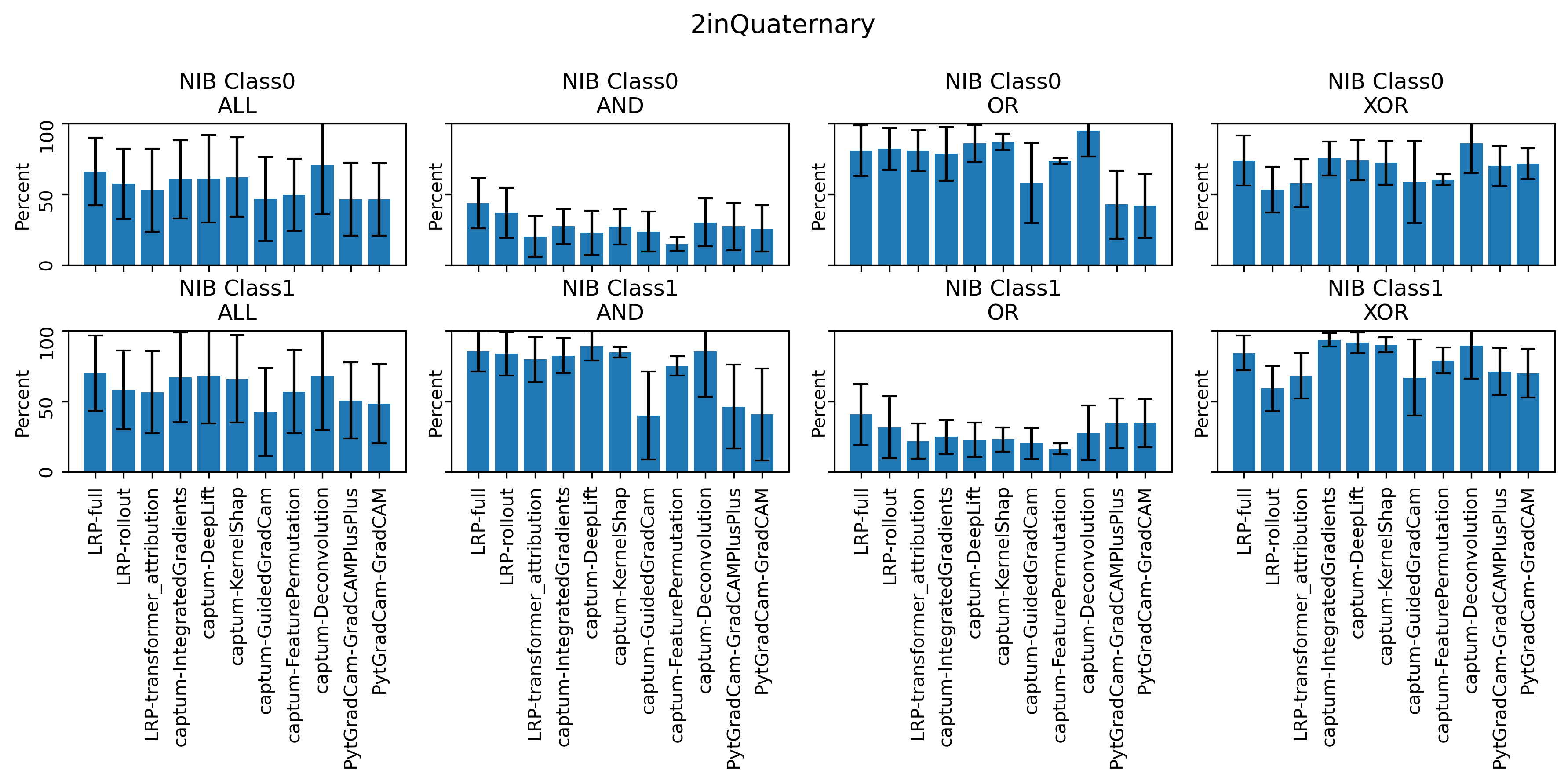}
	\caption{Average NIB per class on test sets of all trained DL-models on the 2inQuaternary dataset.}
	\label{fig:NIB2inQuad}
\end{figure}

\begin{figure}[htb!]
	\centering
	\includegraphics[width=.99\columnwidth]{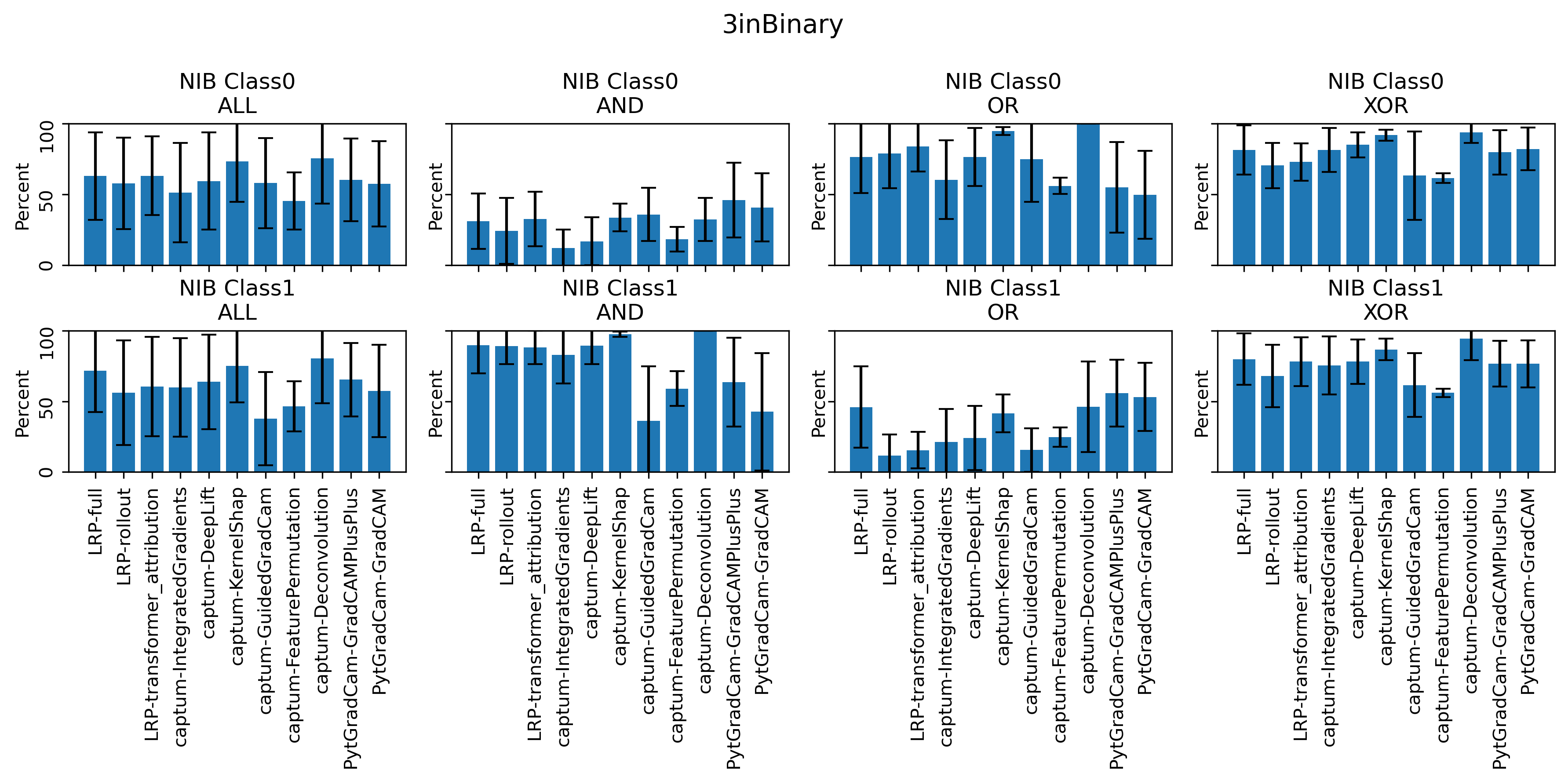}
	\caption{Average NIB per class on test sets of all trained DL-models on the 3inBinary dataset.}
	\label{fig:NIB3inBinary}
\end{figure}

\begin{figure}[htb!]
	\centering
	\includegraphics[width=.99\columnwidth]{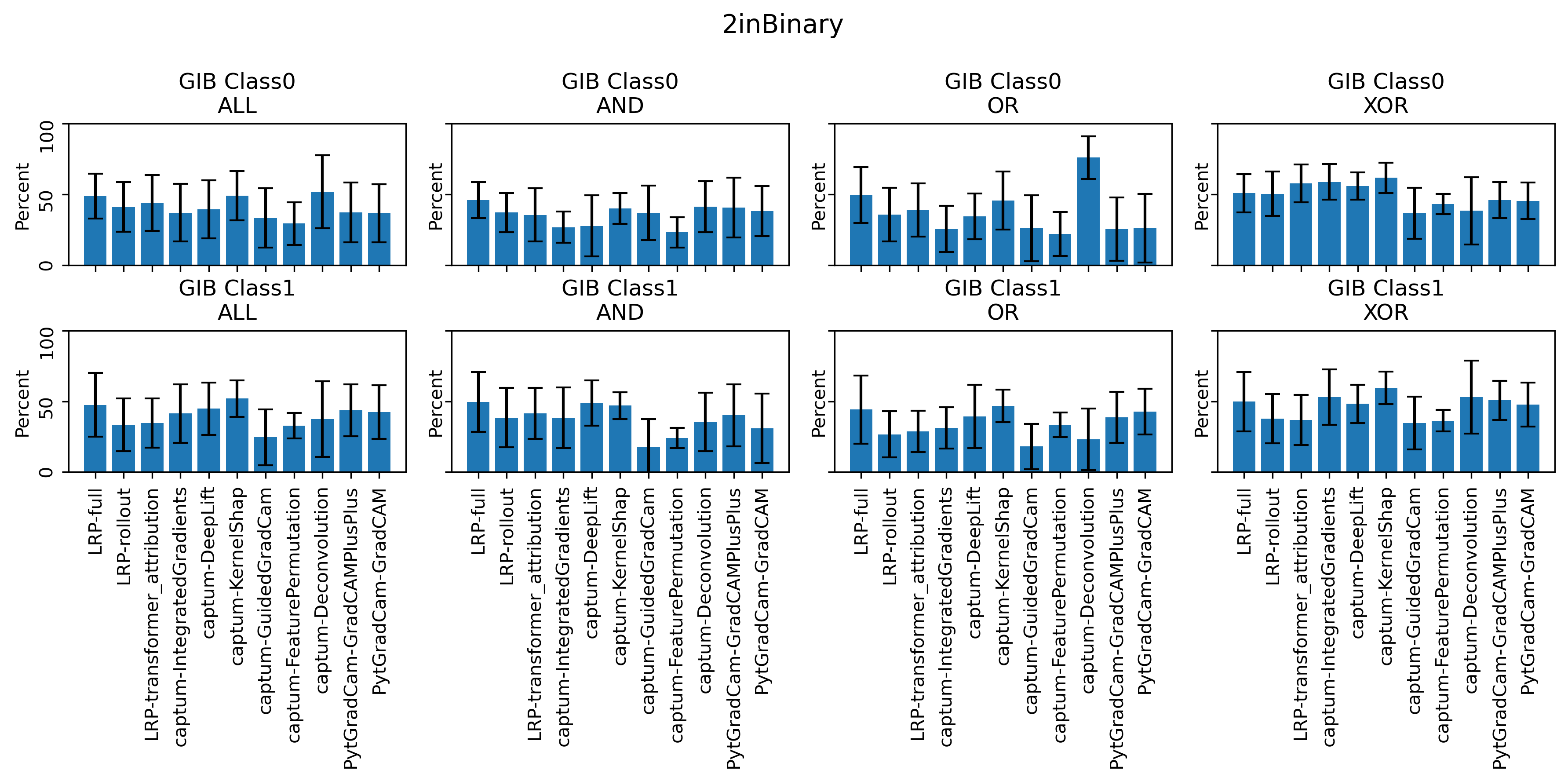}
	\caption{Average GIB per class on test sets of all trained DL-models on the 2inBinary dataset.}
	\label{fig:GIB2inBinary}
\end{figure}

\begin{figure}[htb!]
	\centering
	\includegraphics[width=.99\columnwidth]{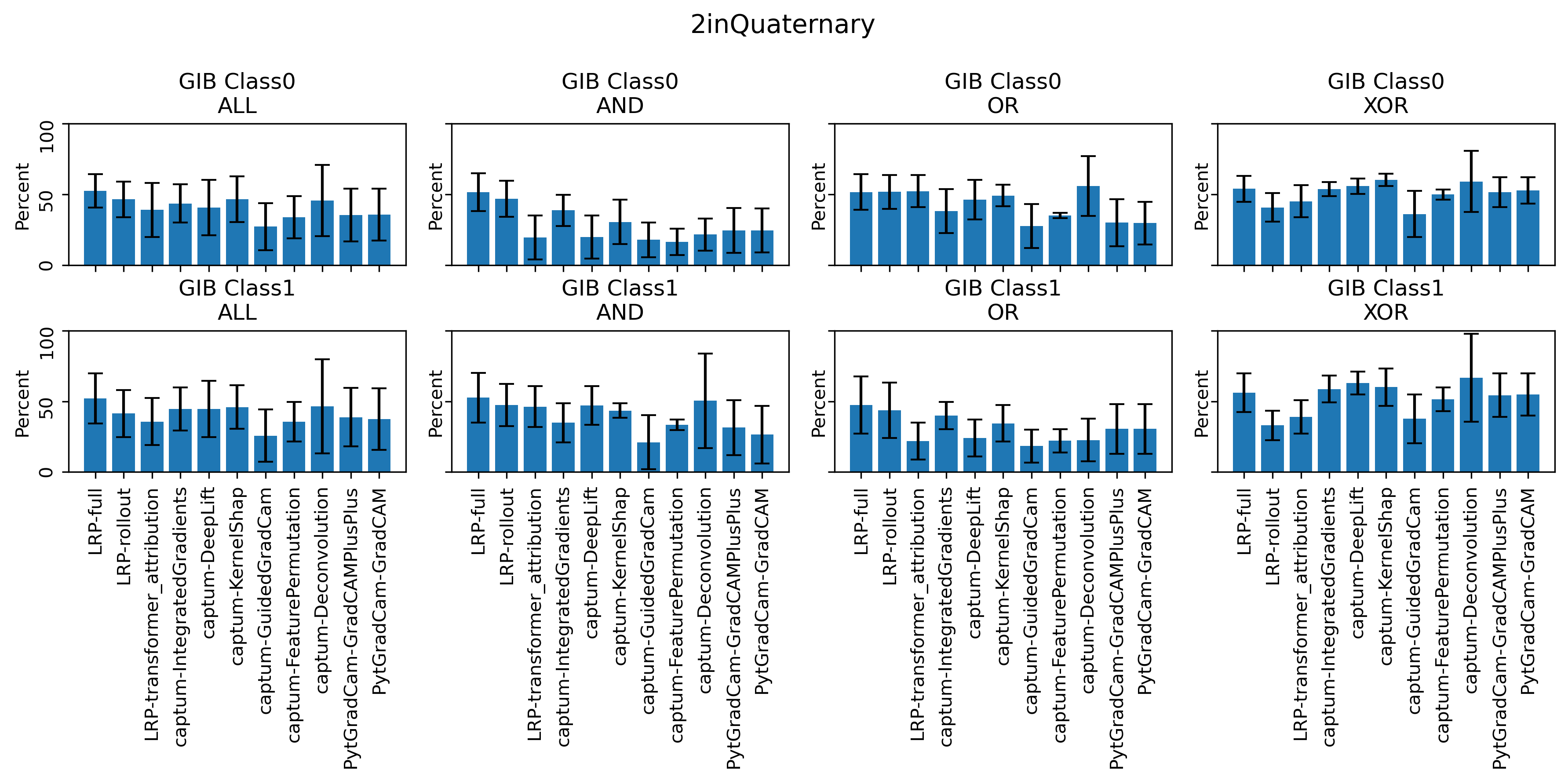}
	\caption{Average GIB per class on test sets of all trained DL-models on the 2inQuaternary dataset.}
	\label{fig:GIB2inQuad}
\end{figure}

\begin{figure}[htb!]
	\centering
	\includegraphics[width=.99\columnwidth]{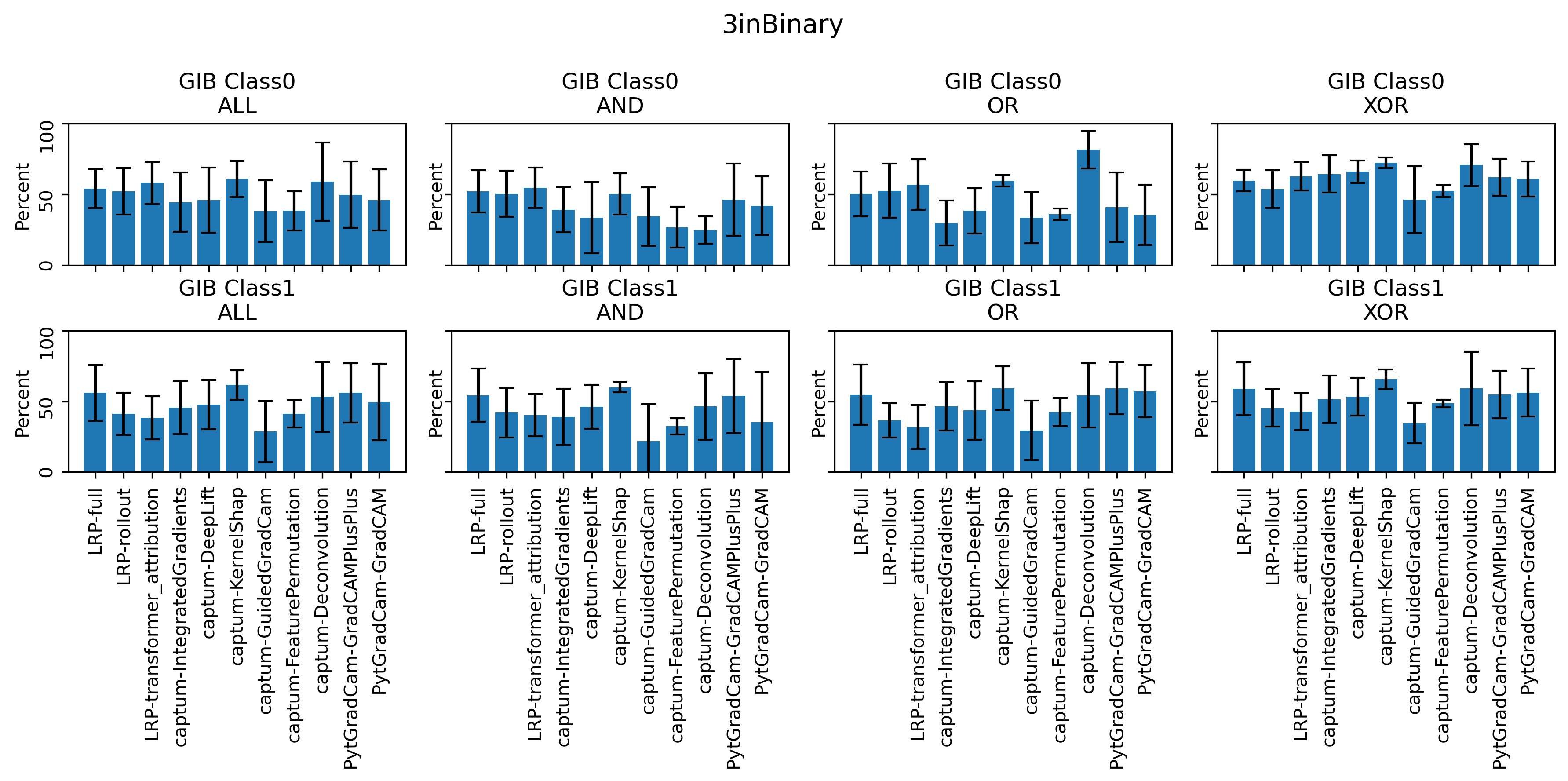}
	\caption{Average GIB per class on test sets of all trained DL-models on the 3inBinary dataset.}
	\label{fig:GIB3inBinary}
\end{figure}

\FloatBarrier
\subsection{Retrained Model Results}
\label{app:retrainedResults}

\begin{figure}[htb!]
	\centering
	\includegraphics[width=.99\columnwidth]{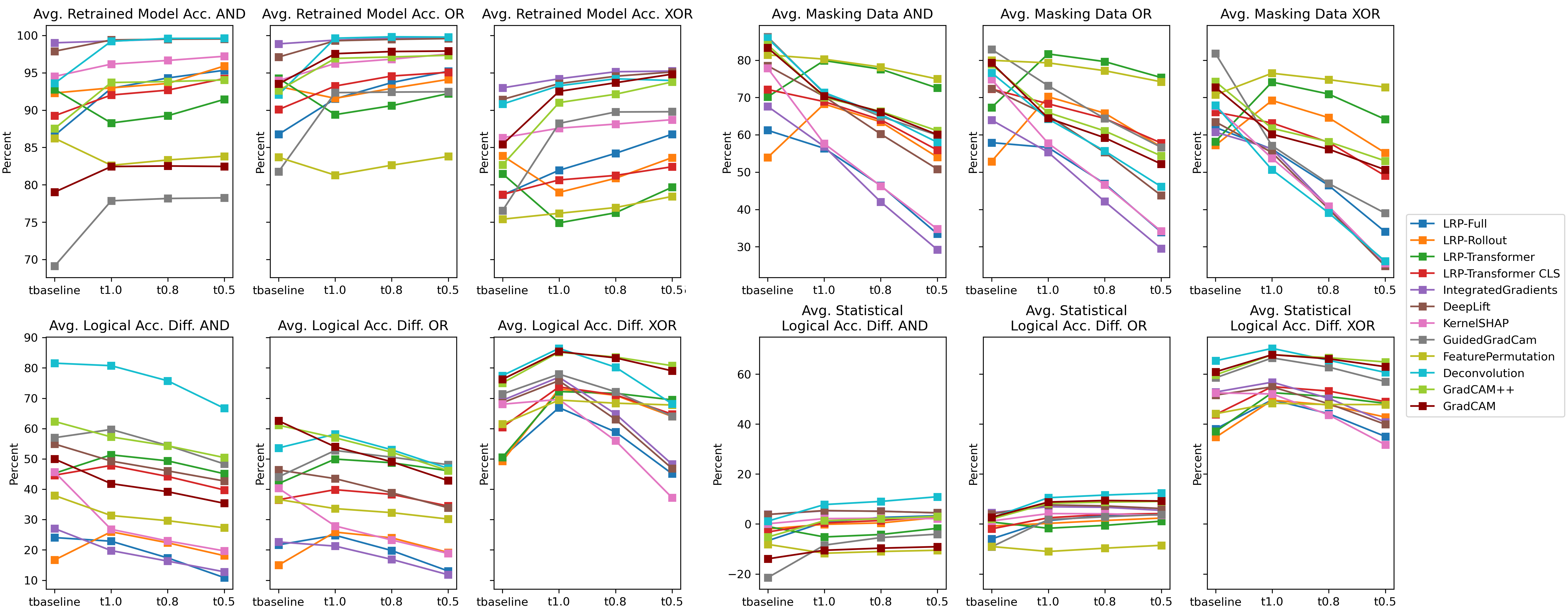}
	\caption{Average retrained model acc., avg masked data, avg. logical acc. difference and avg. statistical logical acc. difference (diff. to retrained model acc.) on the test sets of all trained DL-models.}
	\label{fig:performanceAll}
\end{figure}

\begin{figure}[htb!]
	\centering
	\includegraphics[width=.99\columnwidth]{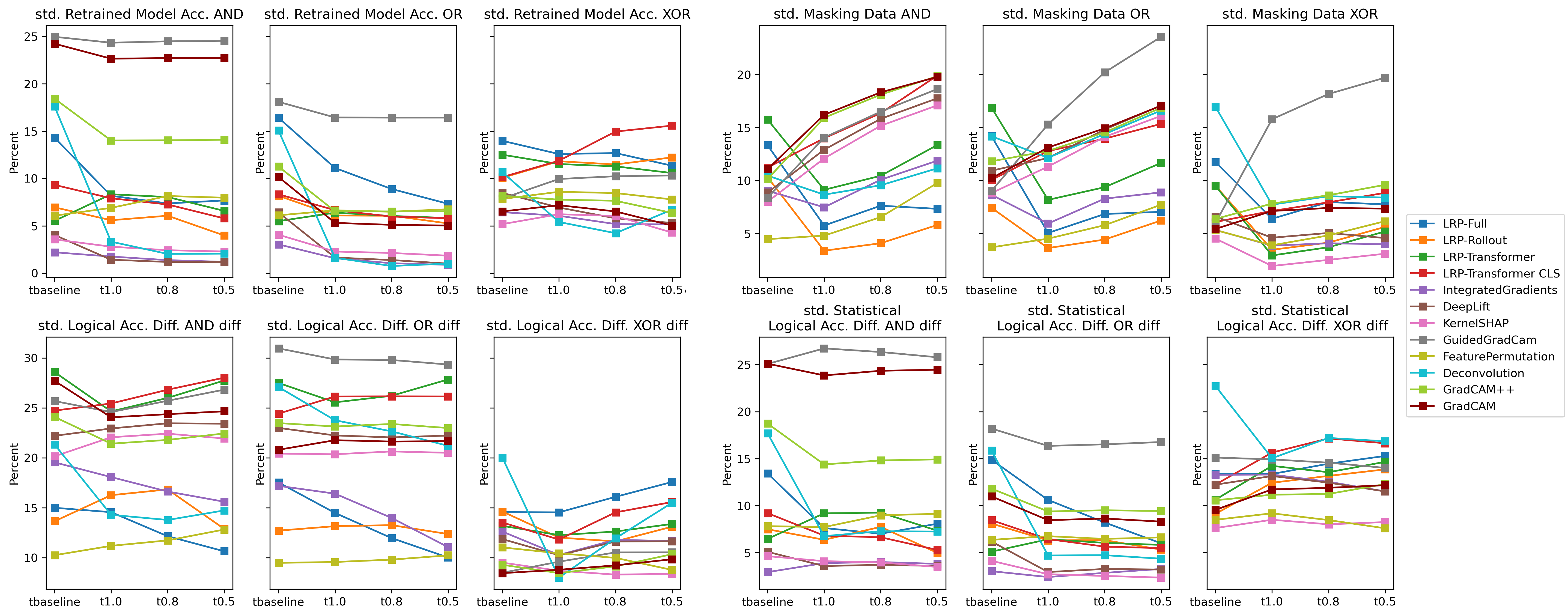}
	\caption{Standard deviation for: Average retrained model acc., avg masked data, avg. logical acc. difference and avg. statistical logical acc. difference (diff. to retrained model acc.) on the test sets of all trained DL-models.}
	\label{fig:performanceAllStd}
\end{figure}

\begin{figure}[htb!]
	\centering
	\includegraphics[width=.99\columnwidth]{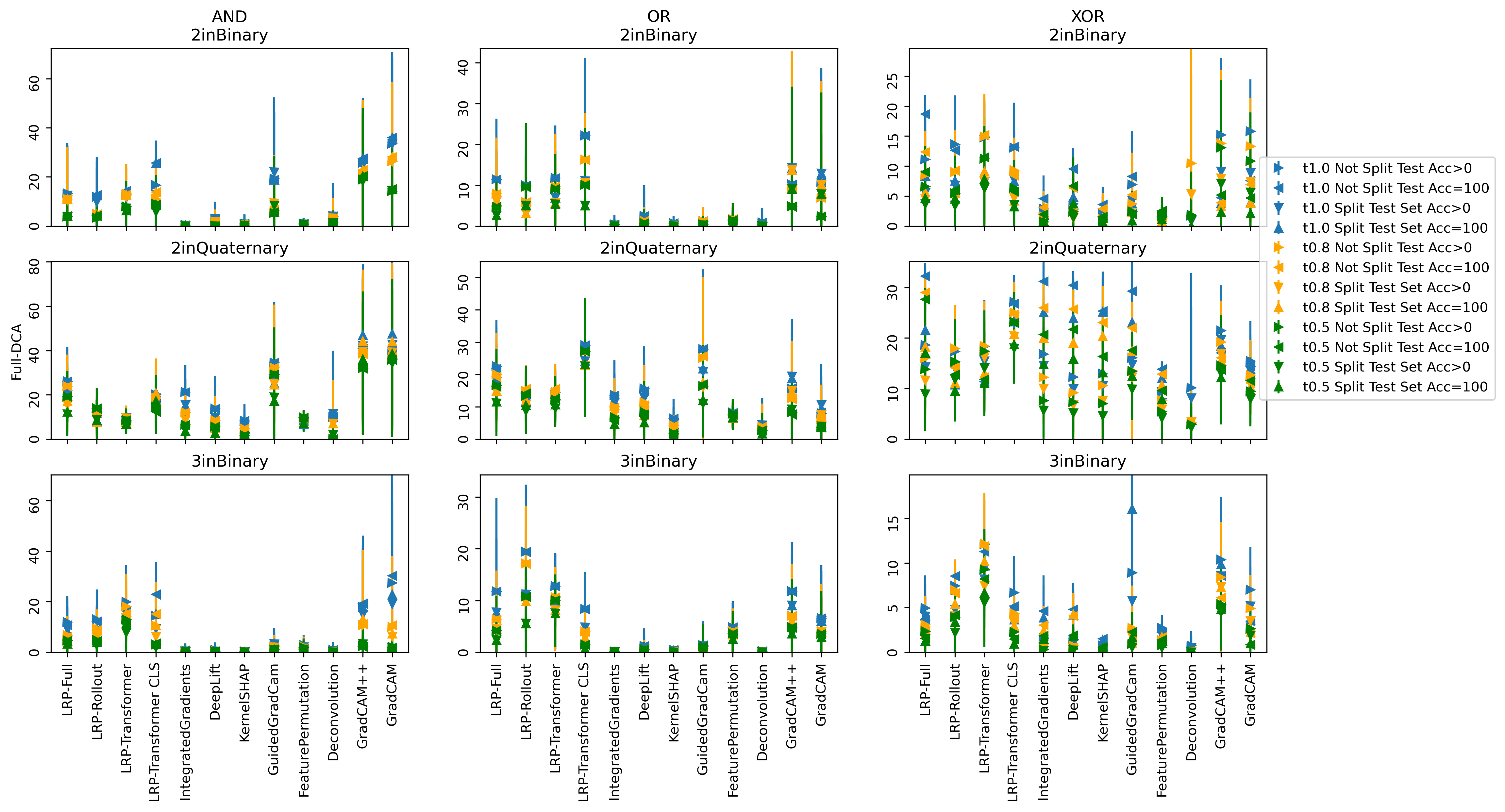}
	\caption{Full DCA, showing how often relevant combinations occur as positive and as negative class after masking, based on 4 different thresholds per saliency method; done on all trained models, split to for four different conditions.}
	\label{fig:FullDCAAll}
\end{figure}

\begin{figure}[htb!]
	\centering
	\includegraphics[width=.99\columnwidth]{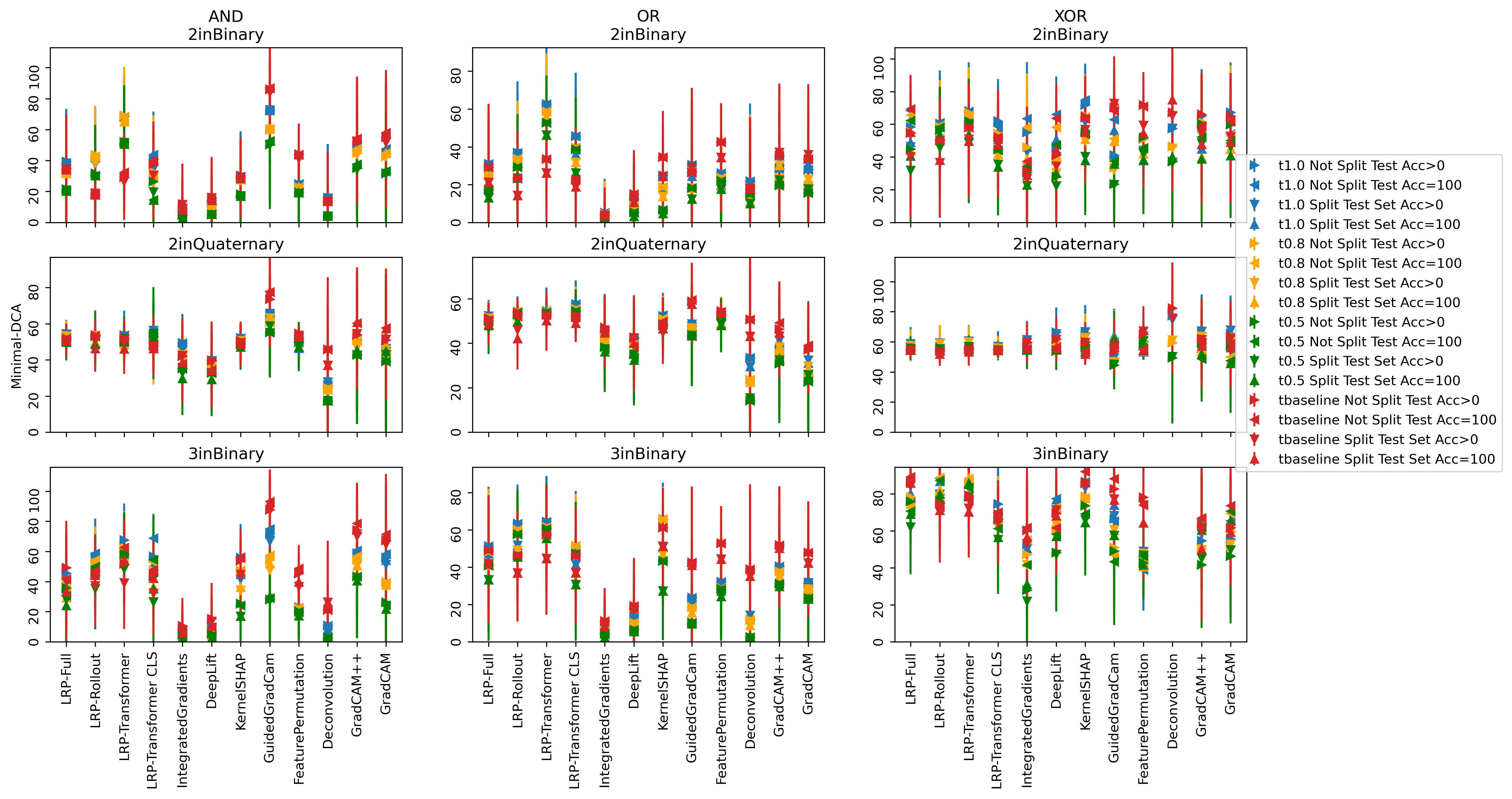}
	\caption{Minimal DCA, showing how often relevant gate combinations occur as positive and as negative class after masking, based on 4 different thresholds per saliency method; done on all trained models, split to for four different conditions.}
	\label{fig:MinDCAAll}
\end{figure}
\end{document}